\ifdefined\pdfobjcompresslevel
\fi
\RequirePackage{fix-cm}
\RequirePackage{scrlfile}
\PreventPackageFromLoading{CJKutf8}
\PassOptionsToPackage{table,dvipsnames}{xcolor}
\PassOptionsToPackage{sort&compress}{natbib}
\documentclass[11pt,a4paper]{jitagent}
\usepackage{graphicx,booktabs,tabularx,array,multirow}
\usepackage{colortbl}
\usepackage{wrapfig}
\usepackage{bm}
\usepackage{caption}
\newcommand{\symbf}[1]{\bm{#1}}
\newcommand{\hflogo}{\raisebox{-0.22\height}{\includegraphics[height=0.95em]{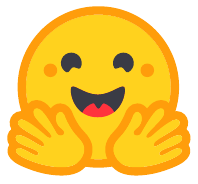}}\hspace{0.3em}}
\usepackage[ruled,vlined,linesnumbered]{algorithm2e}
\tcbuselibrary{most}
\usepackage{placeins}
\usepackage[noabbrev,nameinlink]{cleveref}
\usepackage{enumitem}
\usepackage{fontawesome5}
\usepackage{atria-jit-branding}
\definecolor{atria}{HTML}{2456A6}
\definecolor{atrialight}{HTML}{E9EEF7}
\definecolor{atriablue}{HTML}{2F5E9E}
\definecolor{atriagold}{HTML}{9B6B25}
\hypersetup{colorlinks=true,linkcolor=atria,citecolor=atriablue,urlcolor=atriablue,
  pdfauthor={Guanting Dong, Zhichao Sun, Xinyu Yang, et al.},
  pdftitle={ToolSpace: Scaling Tool-Centric Environment Synthesis for Long-Horizon Agents}}
\setlist[itemize]{leftmargin=1.3em,itemsep=2pt,topsep=4pt}
\setlist[enumerate]{leftmargin=1.5em,itemsep=2pt,topsep=4pt}
\newcolumntype{Y}{>{\raggedright\arraybackslash}X}

\long\def\Description#1{}
\newcommand{\hyp}{\nobreak-\hspace{0pt}}
\newcommand{\amark}[1]{\textsuperscript{\mdseries\fontsize{6.3pt}{6.5pt}\selectfont #1}}
\newcommand{\mlogo}[1]{\raisebox{-0.22\height}{\includegraphics[height=0.95em]{fig/#1.png}}\hspace{0.3em}}
\newcommand{\mlogopdf}[1]{\raisebox{-0.22\height}{\includegraphics[height=0.95em]{fig/#1.pdf}}\hspace{0.3em}}

\definecolor{tspurple}{RGB}{109,40,217}
\DeclareCaptionFont{jitcaptionsmall}{\fontsize{8.2pt}{10.5pt}\selectfont}
\newcommand{\tslogo}{\raisebox{-0.22\height}{\includegraphics[height=1.25em]{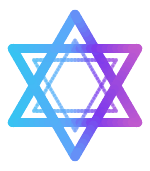}}\hspace{0.3em}}
\preauthor{\par\begingroup\centering\rmfamily\bfseries\fontsize{9.9pt}{12.4pt}\selectfont\spaceskip=0.23em\xspaceskip=0.23em}
\postauthor{\par\endgroup}
\pretitle{\par\begingroup\centering\fontsize{17pt}{20pt}\selectfont\bfseries}
\posttitle{\par\endgroup\noindent{\color{black}\rule{\textwidth}{0.4pt}}\par}
\renewcommand{\jitabstractfont}{\fontsize{9.2pt}{11.7pt}\selectfont}
\title{\raisebox{-4.2pt}{\includegraphics[height=1.18em]{fig/toolspace.pdf}}\hspace{0.14em}\textcolor[HTML]{2447D6}{T}\textcolor[HTML]{3540D7}{o}\textcolor[HTML]{4539D7}{o}\textcolor[HTML]{5632D8}{l}\textcolor[HTML]{662BD9}{S}\textcolor[HTML]{752EDB}{p}\textcolor[HTML]{8339DF}{a}\textcolor[HTML]{9044E2}{c}\textcolor[HTML]{9E4FE6}{e}: Scaling Tool-Centric Environment Synthesis\\ for Long-Horizon Agents}
\author{Guanting Dong\amark{1,$\ast$}, Zhichao Sun\amark{2,$\ast$}, Xinyu Yang\amark{1,$\ast$}, Rui Wang\amark{2}, Xin Deng\amark{2}, Jiajie Jin\amark{1}, Zhixiong Yang\amark{2},\\ Junhao Huang\amark{3}, Han Han\amark{2}, Mingyu Ma\amark{2}, Xiaotao Wang\amark{2}, Yuyang Hu\amark{1}, Xiaoshuai Song\amark{1},\\ Xiaoxi Li\amark{1}, Jiejun Tan\amark{1}, Zhecan Wang\amark{2}, Shihan Dou\amark{2}, Honglin Guo\amark{2}, Zhiheng Xi\amark{2},\\ Tao Gui\amark{2,$\dagger$}, Tao Ji\amark{2}, Qi Zhang\amark{2}, Yutao Zhu\amark{1}, Ji-Rong Wen\amark{1}, Zhicheng Dou\amark{1,$\dagger$}\\[6pt] {\mdseries\fontsize{9.5pt}{11pt}\selectfont \amark{1}Renmin University of China \quad \amark{2}Fudan University \quad \amark{3}Shanghai Jiao Tong University}\\[2.5pt] {\mdseries\fontsize{8.5pt}{10.5pt}\selectfont \amark{$\ast$}Equal contribution.\quad \amark{$\dagger$}Corresponding authors.}}

\renewcommand{\institutionlogos}{%
  \includegraphics[height=1.2cm]{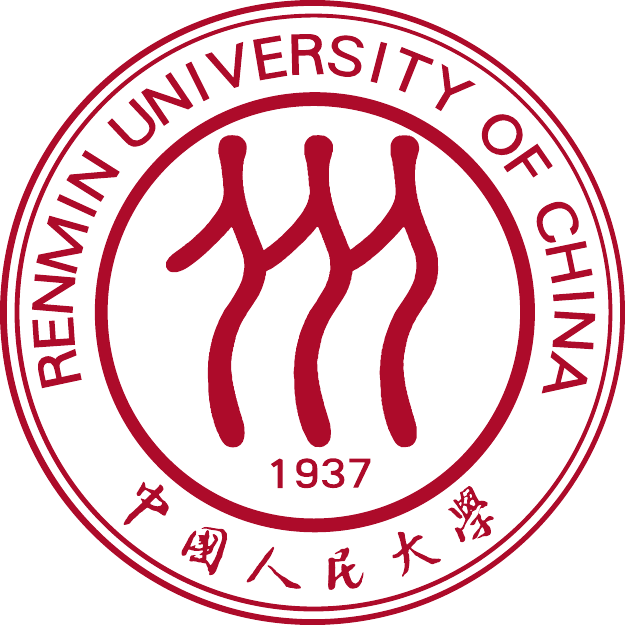}%
  \hspace{0.65em}%
  \includegraphics[height=1.2cm]{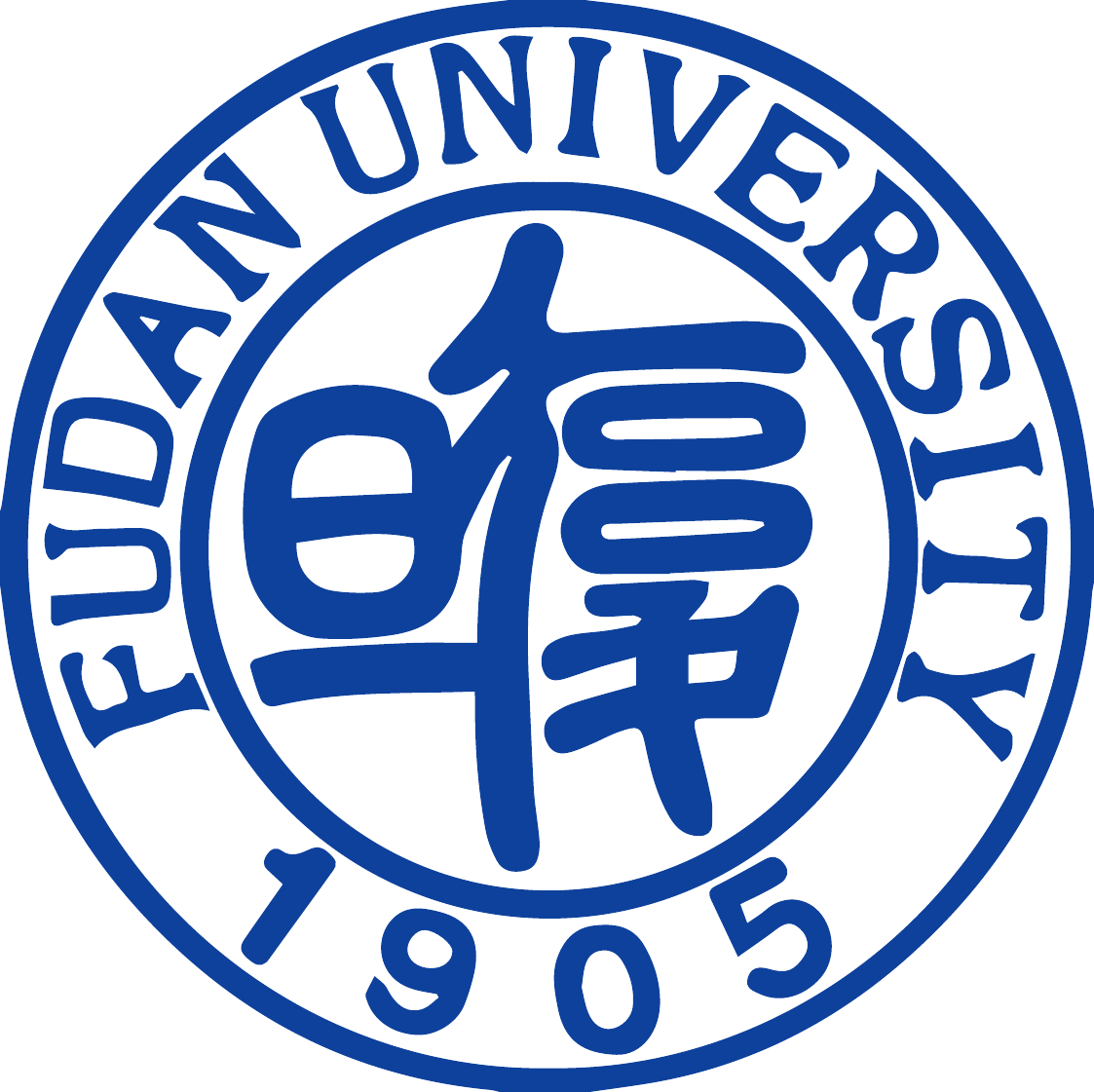}%
  \hspace{0.65em}%
  \includegraphics[height=1.2cm]{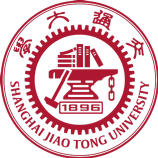}%
}
\renewcommand{\placeinstitutionheader}{%
  \AddToShipoutPictureFG*{%
    \put(\LenToUnit{\dimexpr1in+\oddsidemargin\relax},%
         \LenToUnit{\dimexpr\paperheight-1.2cm+0.8em\relax}){%
      \makebox(0,0)[lt]{\institutionlogos}}%
    \put(\LenToUnit{\dimexpr1in+\oddsidemargin\relax},%
         \LenToUnit{\dimexpr\paperheight-2.5cm\relax}){%
      \makebox(0,0)[lt]{{\normalcolor\rule{\textwidth}{0.4pt}}}}%
  }%
}
\metadata[\textcolor{atriablue}{\faEnvelope}\ Contact]{dongguanting@ruc.edu.cn, dou@ruc.edu.cn}
\metadata[\textcolor{atria}{\faGlobe}\ Project Page]{\url{https://toolspace-agent.github.io/}}
\metadata[\textcolor{atria}{\faGithub}\ Github]{\url{https://github.com/dongguanting/ToolSpace}}
\metadata[\hflogo Models]{\url{https://huggingface.co/collections/dongguanting/toolspace}}
\begin{document}
\placeinstitutionheader
\maketitle

\begin{abstract}
Large language models are deployed as long-horizon agents that compose diverse tools across MCP services into executable workflows. Existing MCP environment-scaling methods are data-centric: they first synthesize a database and scenario, then derive tools and tasks from them. As a result, tool interfaces may not match real ones, tool compositions are not grounded in execution, and tasks rarely span more than one environment. We propose ToolSpace, a tool-centric framework that reverses this order. ToolSpace mines data, state, and retry dependencies from adjacent calls in execution traces of real MCP services and open agent corpora, using deterministic rules rather than LLM judgment, and clusters the resulting weighted tool-composition graph into environments. Databases are synthesized only for state-mutating environments, while query and stateless tools replay recorded returns. ToolSpace then performs state-aware random walks with rollback and decompiles each chain into a natural-language task with assertion-based checks, keeping only samples that replay successfully from the initial state. Applied to AutomationBench and $\tau^3$-Bench Banking, the pipeline produces harder Pro versions on which 7 frontier models drop by up to 38.2 points. Across seven long-horizon tool-use benchmarks, ToolSpace raises Qwen3.6-35B-A3B from 23.0 to 41.0 average score (43.2 with RL) and GLM-5.2 from 44.6 to 56.6, on par with Claude Opus 5. Ablations show that execution-witnessed edges clearly outperform LLM-judged ones, and performance keeps improving with data and environment scale.
\printprojectmetadata
\end{abstract}

\begin{figure}[H]
\centering
\makebox[\linewidth][c]{\includegraphics[width=1.00\linewidth,trim=13 10 16 9,clip]{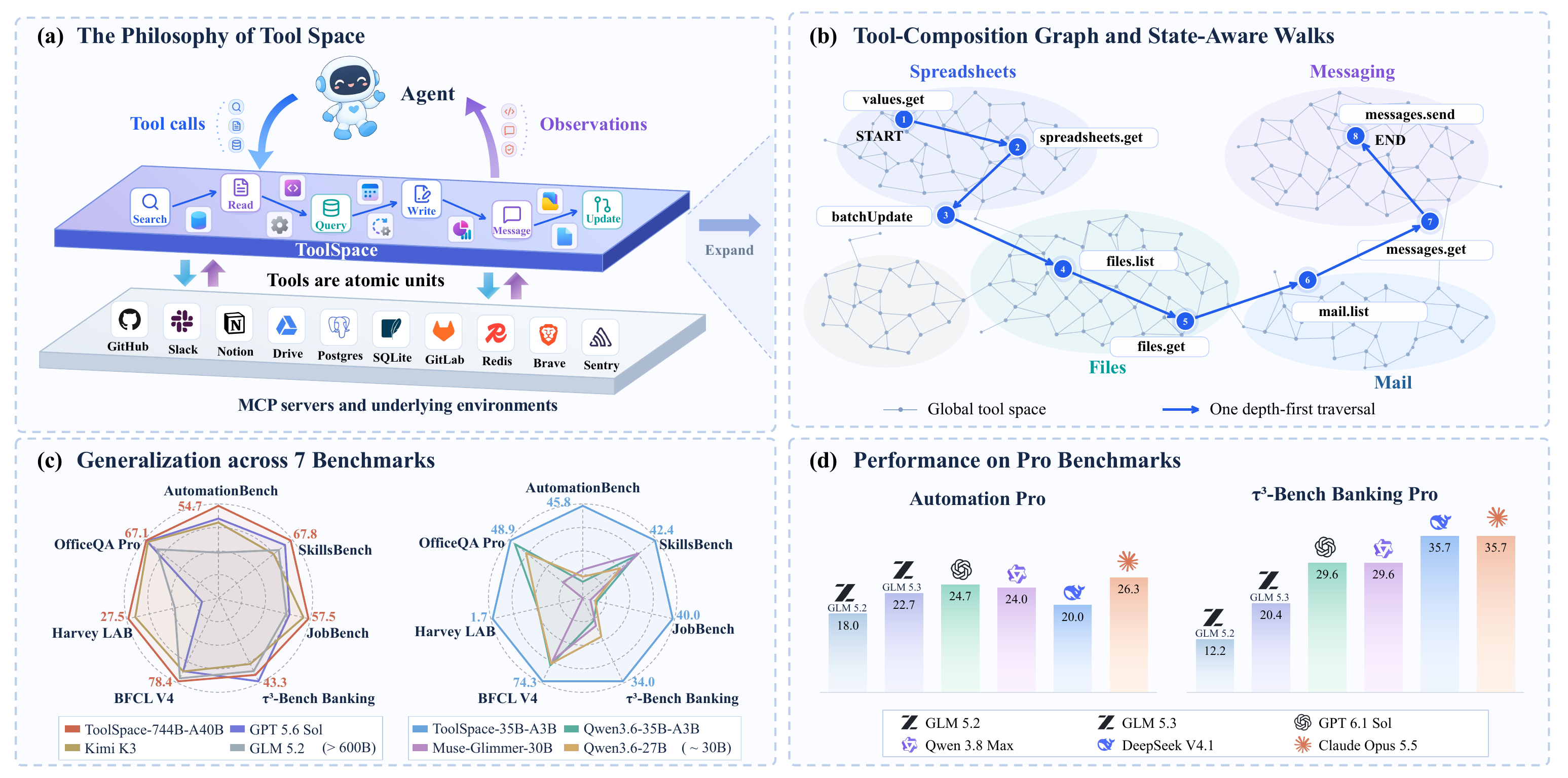}}
\caption{\textbf{Overview of ToolSpace.} (a)~Design Philosophy: a tool space layer between agents and MCP environments. (b)~The tool-composition graph with a state-aware walk. (c)~Generalization across 7 benchmarks. (d)~Performance on two Pro benchmarks.}
\label{fig:intro}
\end{figure}

\section{Introduction}
\label{sec:intro}

Recent advances have extended large language models (LLMs) from single-turn chatbots to collaborative agents that complete long\hyp{}horizon tasks in dynamic environments~\citep{react,toolformer,toolllm,gorilla}. Such agents must not only master a scalable tool set, but also orchestrate diverse tools into executable long\hyp{}horizon chains. Because tools are the atomic units of agent execution, how tools are organized and composed at scale has become a foundational question for expanding agent capability~\citep{apibank,restgpt,toucan,mcp}.

To scale tool use, recent work follows two main routes. One family, exemplified by Simia~\citep{simia} and GenEnv~\citep{genenv}, treats the LLM as an implicit textual world model and generates simulated feedback from tool specifications and interaction histories. These methods are inexpensive, but their feedback is prone to hallucination and state consistency is difficult to guarantee. The second family explicitly reproduces interaction logic by synthesizing executable tools together with environment databases: it extracts or constructs an underlying database from a real service theme, generates matching tools, and synthesizes tasks with verifiable outcomes~\citep{envscaler,awm,dong2026agent,envscaling}. Later systems further evolve whole interaction stacks or repair environments from execution feedback~\citep{weft,workrover,agentreflex}.

Despite these advances, existing paradigms still follow \emph{Data-Centric Environment Synthesis}: synthesize the underlying database and thematic scenario first, and only then derive tools and tasks. This ordering has three inherent limitations:
\begin{itemize}[leftmargin=1.25em,itemsep=1pt,topsep=2pt]
  \item \textbf{Tool--functionality mismatch.} Executable tools are synthesized from the environment database and theme description, so their parameter logic easily diverges from real tool interface definitions. Agents consequently learn a synthetic tool protocol rather than operations transferable to a real tool ecosystem.
  \item \textbf{Inauthentic tool orchestration.} Existing work builds tool-link graphs and synthesizes tasks through random linking or shallow LLM judgments, so the resulting links lack authenticity. Real tool services are moreover unstable and noisy, and their logical relations are hard to establish subjectively.
  \item \textbf{Lack of cross\hyp{}environment task synthesis.} Current task synthesis relies on the data and rules of a single environment, whereas real-world tasks routinely span several environments. Because tools are locked inside their own sandboxes, cross\hyp{}environment composition has no explicit reachability relation, and long\hyp{}horizon complexity cannot grow naturally.
\end{itemize}

Together, these limitations share a single root cause: Data-Centric Environment Synthesis presumes a realistically synthesized environment and derives tool composition from it, reversing the dependency that governs real agent execution. How to synthesize agent training data that preserves real tool orchestration across diverse environments therefore remains an open challenge.

Our starting point is that, \textbf{\emph{within agent execution, tools are the atomic units of capability growth, not static environment data}}. As illustrated in Figure~\ref{fig:intro}(a)--(b), a \emph{Tool Space} that has not yet been explicitly modeled always lies between an agent and its underlying environment databases, and the relations among tools naturally connect an extensible environment ecosystem. A natural path is therefore to model the real connectivity and state transfer among tools, and to build a cross\hyp{}environment tool-graph layer on top of these relations. The complexity of synthesized tasks then arises from expanding the real tool-composition space, rather than from the data scale of any single environment.

Following this insight, we propose \textbf{ToolSpace}, a tool-centric framework for cross\hyp{}environment agent data synthesis. Unlike Data-Centric Environment Synthesis, ToolSpace does not mine a new environment ecosystem from scratch; it follows a \emph{Trace-to-ENV} principle~\citep{trace2env,terminaluniverse}: it first reconstructs a scalable, explorable tool space from execution traces, then synthesizes long\hyp{}horizon agent tasks across the tool and environment clusters connected in this space. ToolSpace consists of two stages:

\begin{itemize}[leftmargin=1.25em,itemsep=1pt,topsep=2pt]
  \item \textbf{Tool Space Exploration.} We extract tool calls from real-world execution traces and mine data, state, and retry links with deterministic rules, building a scalable tool-composition graph. Tightly coupled tools are clustered into environment clusters; an underlying database is synthesized on demand only for operation environments, while original tool returns are reused directly otherwise. This preserves tool-composition relationships while substantially reducing environment synthesis cost.
  \item \textbf{Cross\hyp{}Environment Verifiable Task Synthesis.} Over the ToolSpace graph, State-Aware Random Walks with Rollback Sampling sample state-compatible successors by edge weight and execute them step by step, retaining only chains that execute end to end under runtime-observed state transitions. Each retained walk is then decompiled into a natural-language task with assertion-based verification scripts, yielding training samples that can be replayed and audited across environments.
\end{itemize}

To validate ToolSpace's ability to flexibly synthesize complex cross\hyp{}environment tasks, we anchor the full pipeline to AutomationBench~\citep{automationbench} and $\tau^3$-Bench~Banking~\citep{tauknowledge}, two long\hyp{}horizon agent benchmarks and extend their difficulty, producing \textbf{AutomationBench Pro} and \textbf{$\tau^3$-Bench~Banking Pro}. The upgraded versions exceed the standard ones in task difficulty and number of environments crossed. On AutomationBench Pro, seven frontier models drop by up to 38.2 points relative to the standard version (Figure~\ref{fig:intro}).

In summary, our main contributions are as follows:
\begin{itemize}[leftmargin=1.25em,itemsep=1pt,topsep=2pt]
  \item We identify limits of data-centric synthesis: reverse-synthesized tools mismatch real tool semantics, tool graphs lack execution evidence, and tasks are confined to single-environment complexity; and we then propose \emph{Tool-Centric \mbox{Environment Synthesis}}.
  \item We propose ToolSpace, which reconstructs a scalable, explorable tool-composition space from real execution traces and synthesizes long\hyp{}horizon, verifiable agent tasks across diverse tool and environment combinations within this space.
  \item We show that ToolSpace plugs into long\hyp{}horizon tool\hyp{}use benchmarks, producing AutomationBench Pro and $\tau^3$-Bench Banking Pro, which exceed their standard versions in difficulty, tool-call turns, and cross\hyp{}environment breadth.
  \item \begingroup Across seven long\hyp{}horizon tool\hyp{}use benchmarks, experiments demonstrate clear advantages across model scales (8B--744B). ToolSpace-744B-A40B surpasses GPT 5.6 Sol in average score (56.6 vs.\ 48.9) and matches Claude Opus 5. Further analyses reveal clear scaling relationships of data scale, environment scale, and task diversity with agent performance.\par\endgroup
\end{itemize}

\section{Related Work}
\label{sec:related}

\begingroup
\textbf{Agent Environment Scaling.} Agent Environment Scaling has become an important foundation for the evolution of agent capabilities, by scaling the \emph{environments} in which agents act. Simulator-style methods, exemplified by Simia~\citep{simia} and GenEnv~\citep{genenv}, use an LLM as a textual world model to emulate tool feedback: cheap to scale, but prone to hallucination and state inconsistency. Executable\hyp{}environment methods instead synthesize databases, tools, and verifiable tasks programmatically~\citep{envscaler,awm,dong2026agent,envscaling}, with later variants that evolve environments or whole interaction systems~\citep{workrover,weft} or repair them from execution feedback~\citep{agentreflex}. In all of these, the unit of scaling remains the environment (or the surrounding harness), so which tools may compose is fixed before any task is drawn. ToolSpace is complementary: it recovers tool-composition structure witnessed by executions and synthesizes databases only on demand, so the same machinery can \emph{upgrade} benchmarks.
\par\endgroup

\textbf{Graph-Based Agentic Data Synthesis.} Building agent data with intrinsic dependencies on structured graphs is a promising direction. For search and knowledge agents, prior work builds knowledge graphs or formalizes information seeking tasks to generate hard questions and trajectories~\citep{webdancer,webshaper}. For general tool use, function-signature or API-dependency graphs are walked to produce multi-turn calls and verified blueprints~\citep{magnet,apigenmt,dong2026agent}; similar recipes reverse-synthesize trajectories for multimodal GUI and web agents~\citep{tongui}. These chains then feed SFT, agentic RL, and reflection-style training~\citep{react,shao2024deepseekmath,agentreflex,reflexion}. ToolSpace differs in where the graph comes from and what it guarantees: edges are mined by deterministic rules from adjacent real calls~\citep{toolllm,toucan,sweagent,openhands} across environments, typed as data, state, or retry and weighted by witnessed frequency; a walk enters training only after replay from $s_0$ passes state, trace, and output checks. ToolSpace is the framework to make witnessed tool composition the single primitive behind cross\hyp{}environment tasks and benchmarks.

\section{Preliminaries and Problem Formulation}
\label{sec:prelim}

We formalize tool-use tasks, data-centric environment synthesis, and the
tool-centric synthesis problem addressed by ToolSpace.

\textbf{Tool-use tasks.}
Let $\mathcal{F}$ denote a tool space, where each tool
$f\in\mathcal{F}$ has input and output schemas $\alpha_f$ and $\beta_f$.
An environment $e=(\mathcal{D},\mathcal{F}_e)$ pairs a persistent
database $\mathcal{D}$ with a tool subset
$\mathcal{F}_e\subseteq\mathcal{F}$ and induces a state space
$\mathcal{S}_e$ with snapshots $s_t$. Given a natural-language task $q$,
an agent policy $\pi_\theta$ produces
\begin{equation}
\label{eq:traj}
\tau=(a_1,o_1,\ldots,a_L,o_L),\qquad
a_t=\pi_\theta(q,h_t),\quad
o_t=f_t(x_t;s_{t-1}),
\end{equation}
where $h_t$ is the interaction history, $f_t$ is the tool invoked with
arguments $x_t$, $o_t$ is its observation, $L$ is the number of tool
calls, and $s_t$ is the environment state after step $t$. A verifier
$V$ maps $(q,\tau,s_L)$ to $\{0,1\}$, or to a graded assertion score.

\textbf{Data-centric environment synthesis.}
Prior work samples a theme $m$, first synthesizes an underlying
database $\mathcal{D}\sim p(\mathcal{D}\mid m)$, then synthesizes
derived tools $\mathcal{F}_e\sim p(\mathcal{F}_e\mid \mathcal{D},m)$
conditioned jointly on the theme and the database, and further
synthesizes tasks $q\sim p(q\mid \mathcal{D},\mathcal{F}_e)$. Because
$\mathcal{F}_e$ is a synthetic tool set conditioned on a fabricated
$\mathcal{D}$, its argument logic need not coincide with any real
interface, and the resulting task synthesis is likewise detached from
real tool-logic relations; cross-environment reachability is undefined,
as tools from different $(\mathcal{D},\mathcal{F}_e)$ pairs share no
typed relation.

\textbf{Tool-centric environment synthesis (our problem).}
Given real execution traces
$\mathcal{M}=\{(f_t,x_t,o_t)\}$, we seek:
(i) a typed tool graph
$\mathcal{G}_{\mathrm{TS}}=(\mathcal{V},\mathcal{E},w)$ whose tool
nodes and inter-tool connections are extracted from and filtered
against $\mathcal{M}$;
(ii) environment clusters aggregated by tool-edge density together with
real tool definitions;
(iii) underlying data synthesized on demand for operation environments;
and (iv) a dataset
$\mathcal{X}_{\mathrm{TS}}=\{(q,V,\omega,s_0)\}$.
Each tool chain $\omega=(v_1,\ldots,v_L)$ must be a walk on
$\mathcal{G}_{\mathrm{TS}}$ satisfying dependency reachability, state
preconditions with respect to $s_t$, and parameter provenance: every
argument must be traceable to $q$, $s_0$, or an earlier observation
$o_{<t}$. It must also be \emph{replay-verified}, such that re-executing
$\omega$ from $s_0$ passes all assertions in $V$. Our objective is to
maximize the long-horizon, cross-environment coverage of
$\mathcal{X}_{\mathrm{TS}}$ subject to replay verification, rather than
the fidelity of any single synthesized database. Under this
tool-centric formulation, tools conform to real tool-interface
definitions, the execution logic of synthesized tasks follows real
tool-connection relations, and databases are synthesized only on
demand, making synthesis efficient and cost-saving.

\section{Method}
\label{sec:method}

\textbf{Overview.} We present \textbf{ToolSpace}, a tool-centric framework for cross\hyp{}environment agent data synthesis (Figure~\ref{fig:framework}). ToolSpace comprises two tightly coupled components, detailed below, and seamlessly raises the difficulty of existing environment benchmarks (\S\ref{sec:pro}):
\begin{itemize}[leftmargin=1.25em,itemsep=1pt,topsep=2pt]
  \item \textbf{Tool Space Exploration} (\S\ref{sec:space}) extracts tool calls from real traces, mines data/state/retry dependencies by rules, builds a tool-composition graph, clusters tools into environments, and synthesizes a database only for operation-bearing clusters.
  \item \textbf{Cross\hyp{}Environment Task Synthesis} (\S\ref{sec:synthesis}) explores executable long\hyp{}horizon chains by State-Aware Random Walks and decompiles them into tasks with verification scripts, retaining only samples that replay from $s_0$ and pass all checks.
\end{itemize}

\subsection{Tool Space Exploration}
\label{sec:space}

\subsubsection{Agent Trajectory Collection}
To obtain diverse, large-scale real execution evidence, we build a unified agent-trajectory corpus $\mathcal{M}$ from two source families.

\textbf{MCP-ecosystem trajectories.} Existing MCP environment scaling work~\citep{toucan,dong2026agent} typically synthesizes task queries over real MCP servers from registries such as Smithery. Because their tool definitions come from real servers, such tasks can be seamlessly re-executed against the live services. We collect a real tool catalog from the registry, keep only tasks whose tool definitions exactly match the real servers, execute them with diverse agents on real MCP services, and record the complete call chains as $\mathcal{M}_{\mathrm{MCP}}$.

\textbf{Web-sourced agent trajectories.}
We additionally use a deep-search agent to collect general tool\hyp{}use
trajectories from the web. We group them by execution provenance, rank
them from high to low by the degree of execution authenticity, and apply
progressively stricter filtering to lower-authenticity data:
\begin{itemize}[leftmargin=1.25em,itemsep=1pt,topsep=2pt]
  \item \textbf{\emph{Real\hyp{}server interaction trajectories (High).}}
  We retain records executed on real services. Toucan-1.5M~\citep{toucan}
  covers 495 MCP servers, 2{,}000+ tools, and 1.5M trajectories, with
  synthetic instructions but real executions. ToolBench/ToolLLM~\citep{toolllm}
  covers 16{,}464 RapidAPI interfaces and $\sim$469K API calls; we remove
  unavailable interfaces.

  \item \textbf{\emph{Real\hyp{}workspace interaction trajectories (Medium).}}
  We collect trajectories executed by SWE-agent~\citep{sweagent} and
  OpenHands~\citep{openhands} in real containers and repositories. We retain
  only explicit tool calls and exclude shell wrappers such as
  \texttt{execute\_bash}.

  \item \textbf{\emph{Synthetic\hyp{}tool trajectories (Low).}}
  We collect trajectories generated in synthetic environments
  ~\citep{envscaler,dong2026agent,agentreflex}. Because of potential tool
  inconsistency, we retain only calls whose tool names and schema formats
  match corresponding real tools.
\end{itemize}

The three groups merge into $\mathcal{M}_{\mathrm{open}}$; all tool calls form
\begin{equation}
\label{eq:corpus}
\mathcal{M}=\mathcal{M}_{\mathrm{MCP}}\cup\mathcal{M}_{\mathrm{open}}=\{(f_t,x_t,o_t)\},
\end{equation}
where $(f_t,x_t,o_t)$ is the $t$-th call of a trajectory: $f_t$ the tool, $x_t$ its inputs, and $o_t$ its result. Calls within a trajectory keep their order.

\subsubsection{Tool-Space Graph Modeling}
\label{sec:graph}
\begin{figure}[t]
\centering
\includegraphics[width=\dimexpr\linewidth-1em\relax]{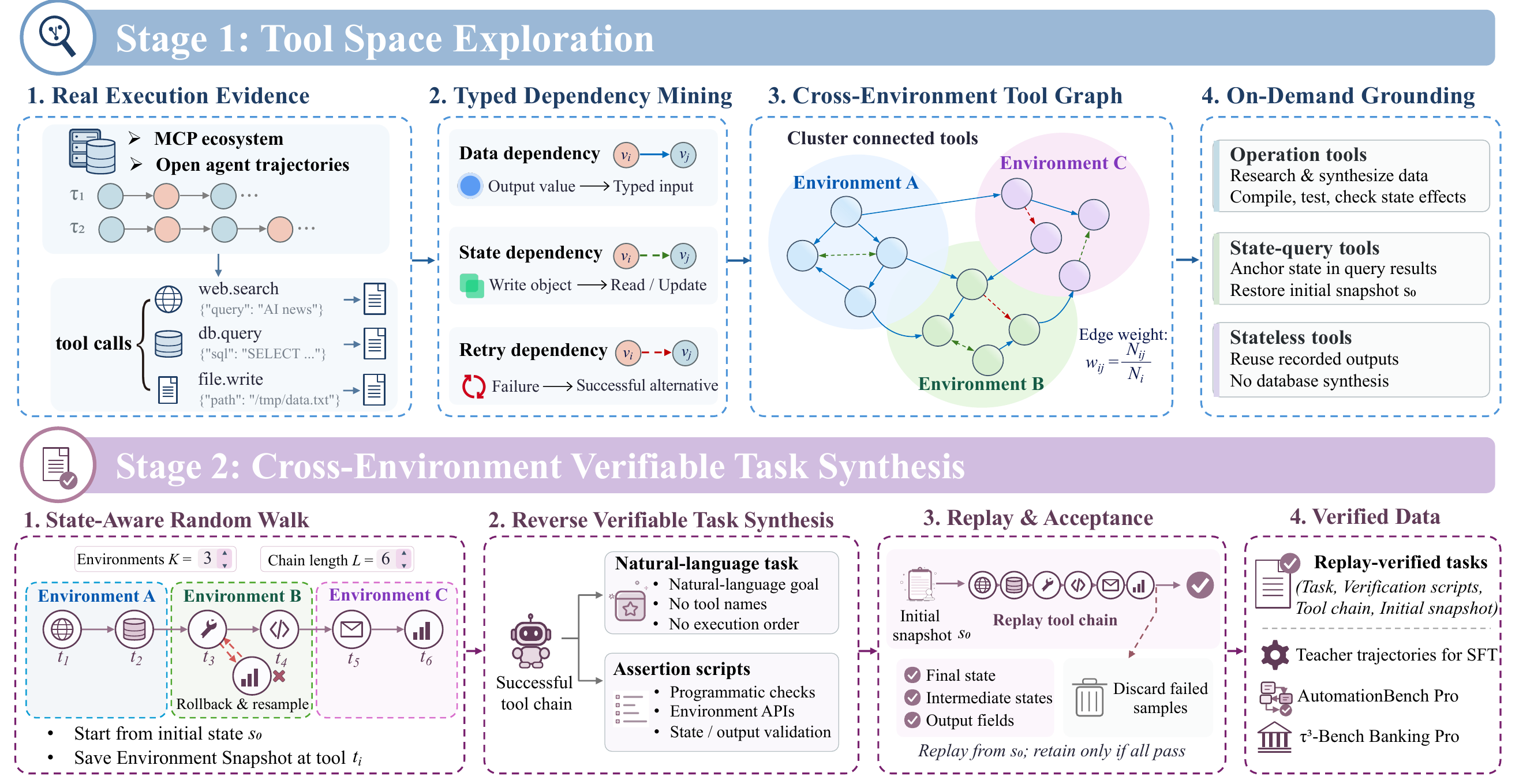}
\Description{Two-stage ToolSpace framework. Stage 1: real execution evidence, typed dependency mining, cross-environment tool graph, and on-demand grounding. Stage 2: State-Aware Random Walks with Rollback Sampling, Reverse Verifiable Task Synthesis, replay and acceptance, and verified SFT data plus Pro benchmarks.}

\captionsetup{font=jitcaptionsmall}
\caption{\textbf{Overview of the ToolSpace framework.} Stage~1 builds the Tool Space from real traces; Stage~2 synthesizes cross-environment tasks by State-Aware Random Walks, retaining only samples that pass replay verification.}
\label{fig:framework}
\end{figure}

To explicitly model these implicit tool relations, we organize $\mathcal{M}$ into a typed tool-composition graph $\mathcal{G}_{\mathrm{TS}}=(\mathcal{V},\mathcal{E},w)$ that can chain different services into one tool world (Figure~\ref{fig:framework}).

\textbf{Nodes.}
For every tool call $f$ occurring in $\mathcal{M}$, we create a node
$v_{f,\tilde{x}}=(n_f,\alpha_f,\beta_f,\tilde{x})$, where $n_f$ is the
tool name, $\alpha_f$ and $\beta_f$ are its input and output schemas,
and $\tilde{x}$ is its normalized argument configuration. Thus, calls
to the same canonical tool with different argument configurations form
distinct nodes, as they may produce different outputs and downstream
dependencies. We write $v_t$ for the node $v_{f_t,\tilde{x}_t}$ instantiated
by the $t$-th call, and $v_i$/$v_j$ for generic nodes below.

\textbf{Relations.} For adjacent calls $(f_t,x_t,o_t)$ and $(f_{t+1},x_{t+1},o_{t+1})$ in a trajectory, we extract three edge types by deterministic rules, \emph{no LLM judgment is used}:
\begin{itemize}[leftmargin=1.25em,itemsep=1pt,topsep=2pt]
  \item \textbf{\emph{Data dependency}} $(v_t\xrightarrow{\mathrm{data}}v_{t+1})$: the input $x_{t+1}$ uses a value from $o_t$ and field types match. Concrete data is passed.
  \item \textbf{\emph{State dependency}} $(v_t\xrightarrow{\mathrm{state}}v_{t+1})$: the first call mutates an environment object and the second call reads or further mutates \emph{the same object}, not merely a coincidentally equal value.
  \item \textbf{\emph{Retry dependency}} $(v_t\xrightarrow{\mathrm{retry}}v_{t+1})$: the first call returns an error, the trajectory switches tools, and the next call succeeds. Such errors are treated as recoverable observations and retained together with the successful retry in the trajectory.
\end{itemize}
Hence, the edge set is defined as
$\mathcal{E}=\mathcal{E}_{\mathrm{data}}\cup
\mathcal{E}_{\mathrm{state}}\cup
\mathcal{E}_{\mathrm{retry}}$.

\textbf{Weights.} To retain recurring tool dependencies rather than tools that merely appear
consecutively, we weight each candidate edge by its observed transition
frequency. Let $N_i$ denote the number of occurrences of node $v_i$ in $\mathcal{M}$
and $N_{ij}$ the number of times $v_j$ immediately follows $v_i$. We define
\[
w(v_i,v_j)=\frac{N_{ij}}{N_i}.
\]
We remove low-frequency edges. For candidate data edges, we additionally
require the downstream arguments to be traceable to the upstream output,
which prevents false links caused by common constants such as
\texttt{limit=10}. Consequently, $\mathcal{G}_{\mathrm{TS}}$ captures tool
transitions during execution rather than merely listing available
tools.

\subsubsection{Agent Environment Data Synthesis}
ToolSpace materializes underlying data only when an environment depends on
persistent state, avoiding the need to synthesize a complete database in
advance. We cluster the weighted undirected projection of
$\mathcal{G}_{\mathrm{TS}}$ by connection density and organize the clusters
using the environment and tool configurations of real MCP services. This
produces $\mathcal{C}=\{c_1,\ldots,c_K\}$ with $K=804$, where each cluster defines an
environment $e_k=(\mathcal{D}_k,\mathcal{F}_k)$ with tool set
$\mathcal{F}_k$ and database $\mathcal{D}_k$. We consider three cases:

\begin{itemize}[leftmargin=1.25em,itemsep=1pt,topsep=2pt]
  \item \textbf{\emph{Operation environments.}}
  These environments contain tools that create, modify, or delete objects
  and therefore require an underlying database. Following
  Agent-World~\citep{dong2026agent}, we derive a theme $m_k$ from each
  cluster and use a research agent $\mathcal{G}_{\mathrm{res}}$, built on
  policy model $\pi_\theta$ and tools $\mathcal{T}_{\mathrm{res}}$, to collect
  and align web data:
  $\mathcal{D}_k^{(0)}=\mathcal{G}_{\mathrm{res}}
  (m_k;\pi_\theta,\mathcal{T}_{\mathrm{res}})$.
  We then iteratively expand it as
  $\mathcal{D}_k^{(n+1)}=\phi(\mathcal{D}_k^{(n)},m_k,
  \mathcal{T}_{\mathrm{res}})$ for $n=0,\ldots,N-1$.
  The environment is retained after unit tests and
  state-transition checks.

  \item \textbf{\emph{State-query environments.}}
  Many state-query tools return partial snapshots of the environment, such
  as \texttt{google\_drive.\allowbreak{}files.\allowbreak{}list} and
  \texttt{gmail.\allowbreak{}users.\allowbreak{}messages.\allowbreak{}get}.
  We use these outputs as anchors to initialize the relevant environment
  state rather than synthesizing a database from scratch. The resulting
  state serves as the initial snapshot $s_0$.

  \item \textbf{\emph{Stateless environments.}}
  These environments contain tools that require no environment-specific
  persistent state, such as web search and fixed-index lookup. For matching
  replay calls, we use cached tool results from $\mathcal{M}$ as execution
  outputs, without issuing live API calls or synthesizing a database.
\end{itemize}

The first two types enter the executable environment system
$\mathcal{E}_{\mathrm{TS}}=\{e_k\}$, while stateless environments use cached
tool results as execution outputs. This design reduces environment-synthesis and API costs and
accelerates subsequent trajectory synthesis.

\begin{algorithm}[!htbp]
\SetAlFnt{\small}
\SetCommentSty{emph}
\SetAlgoLined
\DontPrintSemicolon
\KwIn{trace corpus $\mathcal{M}$, target length $L$, target environments $K$, walk budget}
\KwOut{verified dataset $\mathcal{X}_{\mathrm{TS}}$}
Build $\mathcal{G}_{\mathrm{TS}}=(\mathcal{V},\mathcal{E},w)$ from adjacent calls in $\mathcal{M}$ (data/state/retry rules); prune untraceable or low-weight edges\;
Cluster $\mathcal{G}_{\mathrm{TS}}$ into environments $\{e_k=(\mathcal{D}_k,\mathcal{F}_k)\}$; synthesize $\mathcal{D}_k$ on demand (operation), restore $s_0$ (state-query), or cache returns (stateless)\;
\ForEach{walk attempt}{
  Start at a node bound to $(\mathcal{D}_k,s_0)$; repeatedly sample an admissible neighbor by Eq.~\eqref{eq:walk}, execute it, and update $s_t$\;
  \lIf{execution fails or violates the traversed edge relation}{restore $s_{t-1}$ and resample (rollback)}
  \lIf{length $L$ and $K$ environments are reached}{record $\omega$ with $s_0,\ldots,s_L$}
}
\ForEach{$\omega\in\Omega$}{
  Decompile $\omega$ into task $q$ (no tool names/order) and assertions $V=V_{\mathrm{state}}\cup V_{\mathrm{trace}}\cup V_{\mathrm{out}}$\;
  Replay $\omega$ from $s_0$; \lIf{all assertions pass}{add $(q,V,\omega,s_0)$ to $\mathcal{X}_{\mathrm{TS}}$}
}
Deduplicate $\mathcal{X}_{\mathrm{TS}}$; generate teacher trajectories for retained $q$ and keep those passing $V$\;
\caption{ToolSpace}
\label{alg:toolspace}
\end{algorithm}

\subsection{Cross\hyp{}Environment Task Synthesis}
\label{sec:synthesis}

\subsubsection{State-Aware Random Walks}
To collect long\hyp{}horizon chains that preserve real logical connections, we perform State-Aware Random Walks on $\mathcal{G}_{\mathrm{TS}}$ (Stage~2 of Figure~\ref{fig:framework}). Given a number of crossed environments $K$ and chain length $L$, a walk starts at a tool node bound to a database $\mathcal{D}_k$ and a snapshot $s_0$; $s_t$ denotes the $t$ step environment-state snapshot during execution.

\textbf{Graph Walk.} Let $v_t$ be the current node and consider its neighbors in $\mathcal{E}$. A neighbor is kept only if its next-step inputs can be filled from earlier returns or the current snapshot $s_t$ (dependency reachability, state preconditions, parameter provenance). Among the kept neighbors we sample proportionally to edge weight,
\begin{equation}
\label{eq:walk}
P(v_{t+1}=v\mid v_t)=\frac{w(v_t,v)}{\sum_{u}w(v_t,u)},
\end{equation}
where the sum ranges only over admissible neighbors. Connections that occur more frequently in real traces are therefore more likely to be extended. The walk continues until its length reaches $L$ and its tools cover $K$ environments, yielding a chain $\omega=(v_1,\ldots,v_L)$.

\begin{figure}[t]
\centering
\includegraphics[width=\linewidth]{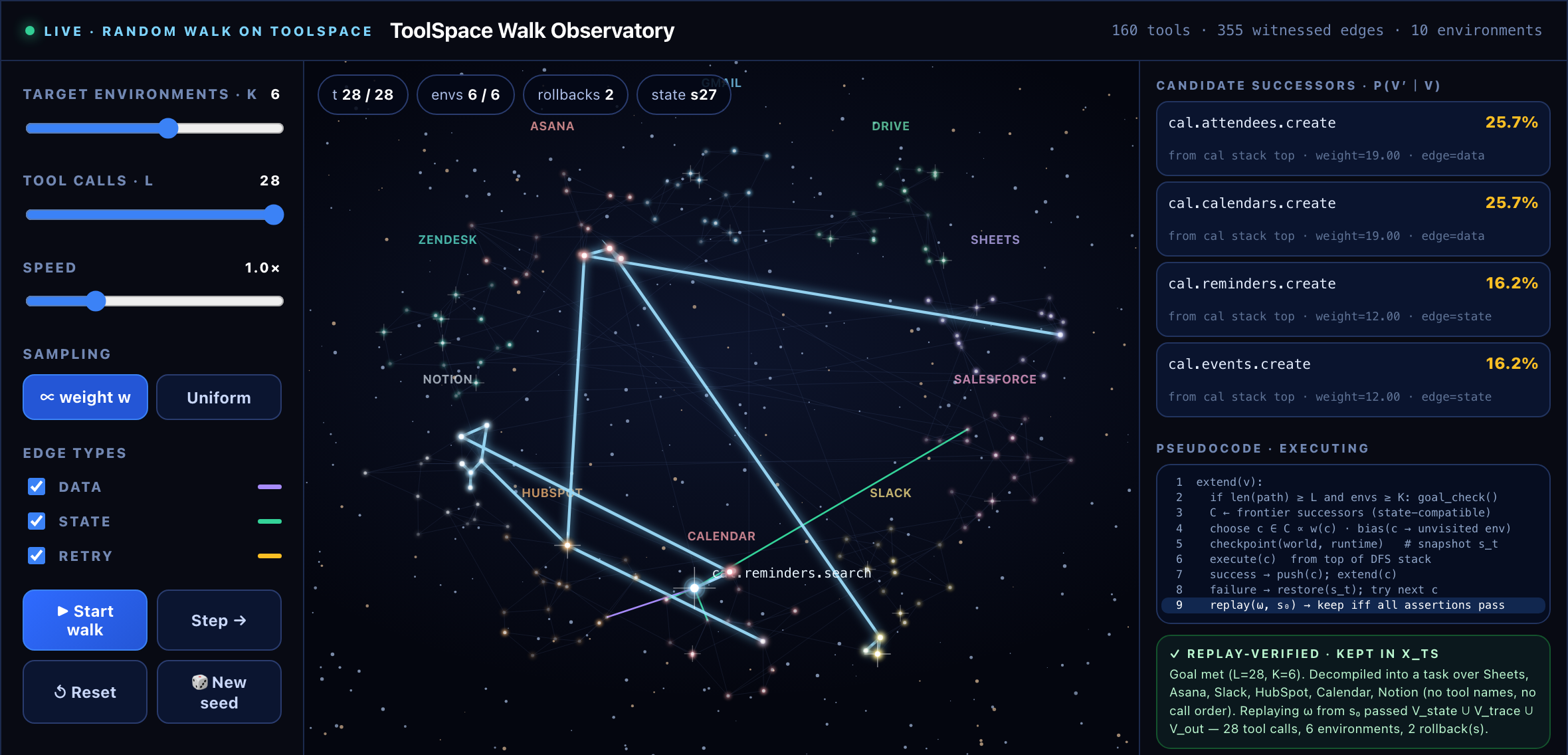}
\caption{\textbf{The Interactive demo of state-aware random walking.}
An example walk from the interactive demo: 28 tool calls across 6 environments, completed after 2 rollbacks and kept after replay verification.
Candidate successors and their sampling probabilities $P(v'\mid v)$ are shown on the right. Demo: \url{https://toolspace-agent.github.io/\#demo}}
\label{fig:walkdemo}
\end{figure}

\textbf{Rollback Sampling.} If a selected tool fails, or the state after execution is inconsistent with the data/state relation of the traversed edge, the step is discarded, the environment is restored to $s_{t-1}$, and another neighbor of $v_{t-1}$ is resampled. Chains that still cannot proceed after repeated rollbacks are discarded. The retained walk set is $\Omega=\{\omega\}$, each walk carrying $s_0$ and snapshots $s_1,\ldots,s_L$. Finally, each retained chain is replayed from $s_0$ (Replay \& Acceptance in Figure~\ref{fig:framework}): re-executing its tools against the recorded snapshots guarantees state continuity along the chain and that every tool remains executable. Figure~\ref{fig:walkdemo} shows one completed walk produced by this procedure in the interactive demo on our project page (\url{https://toolspace-agent.github.io/#demo}).

\subsubsection{Reverse Verifiable Task Synthesis}
To turn executable chains into auditable tasks, we reverse-synthesize each $\omega\in\Omega$. A language model writes a natural-language task $q$ from the chain and its state changes; $q$ contains no tool names and no call order. Instead of relying on a reference answer, we generate a verification set for each chain,
\begin{equation}
\label{eq:verify}
V=V_{\mathrm{state}}\cup V_{\mathrm{trace}}\cup V_{\mathrm{out}},
\end{equation}
where $V_{\mathrm{state}}$ checks the final state $s_L$, $V_{\mathrm{trace}}$ checks key intermediate snapshots, and $V_{\mathrm{out}}$ checks fields required to appear in returned results (here $V$ denotes verification, distinct from the node set $\mathcal{V}$). A sample is retained only if re-executing $\omega$ from $s_0$ passes \emph{all} checks in $V$. Difficulty is raised via $L$, target $K$, and the share of cross\hyp{}environment edges. The dataset is $\mathcal{X}_{\mathrm{TS}}=\{(q,V,\omega,s_0)\}$: task, verification script, tool chain, and snapshot. 
We present the complete ToolSpace algorithm flowchart in Algorithm~\ref{alg:toolspace}.

\subsection{Detailed Statistics of ToolSpace}
\label{app:toolspace-statistics}
\begin{figure}[t]
\centering
\includegraphics[width=\dimexpr\textwidth-1em\relax]{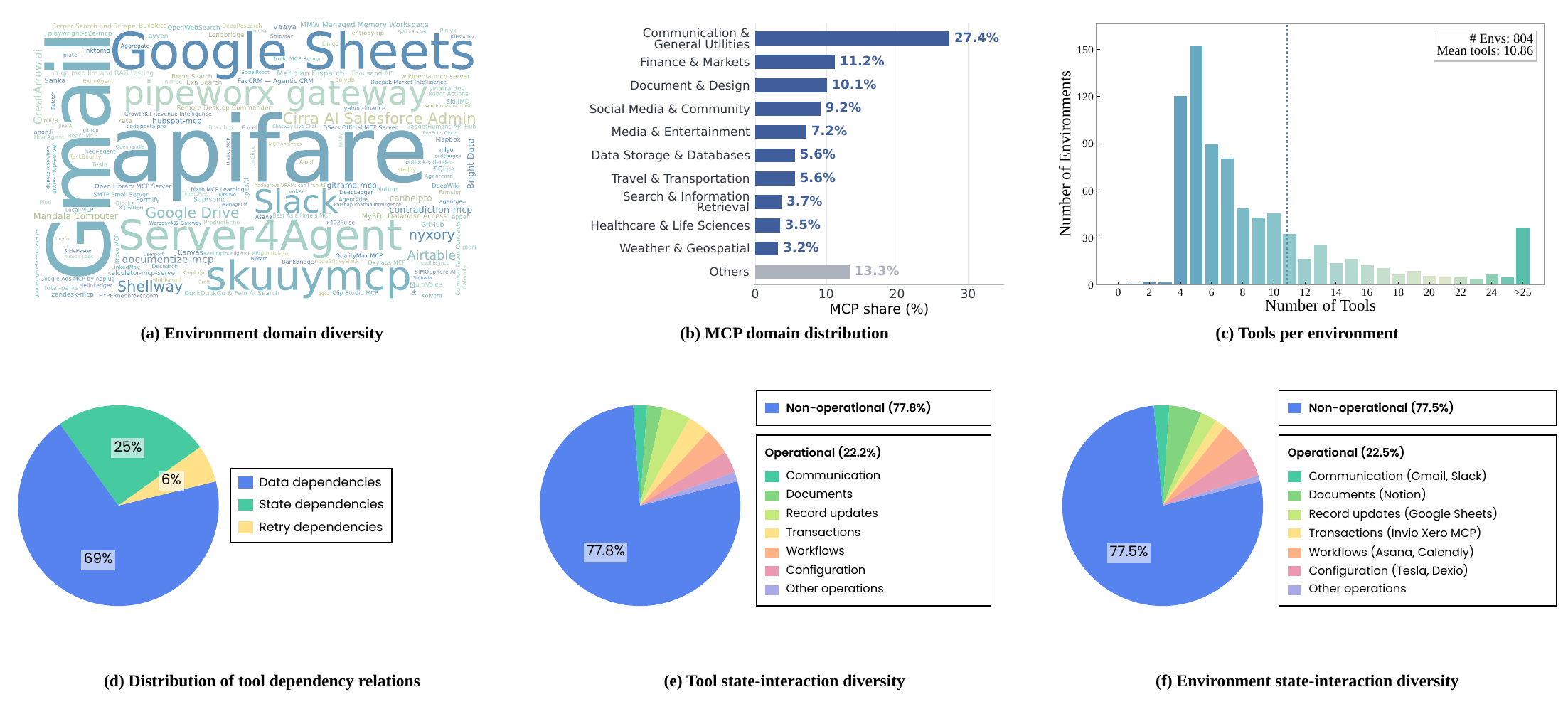}
\Description{Six panels showing real service names, MCP domain distribution across 804 environments, tools per environment, data/state/retry dependency proportions, operational and non-operational tools, and operational and non-operational environments. The domain chart shows the ten largest domains plus Others. The capacity histogram groups values above 25 in the last bin.}

\captionsetup{font=jitcaptionsmall}
\caption{\textbf{ToolSpace graph statistics.}
(a) Environment-name word cloud.
(b) Environment distribution across MCP domains.
(c) Number of tools per environment.
(d) Proportions of data, state, and retry dependencies.
(e) Tool distribution by operational category.
(f) Environment distribution by operational category.}
\label{fig:toolspace-inventory-statistics}
\end{figure}

\begin{figure}[t]
\centering
\includegraphics[width=\dimexpr\textwidth-1em\relax]{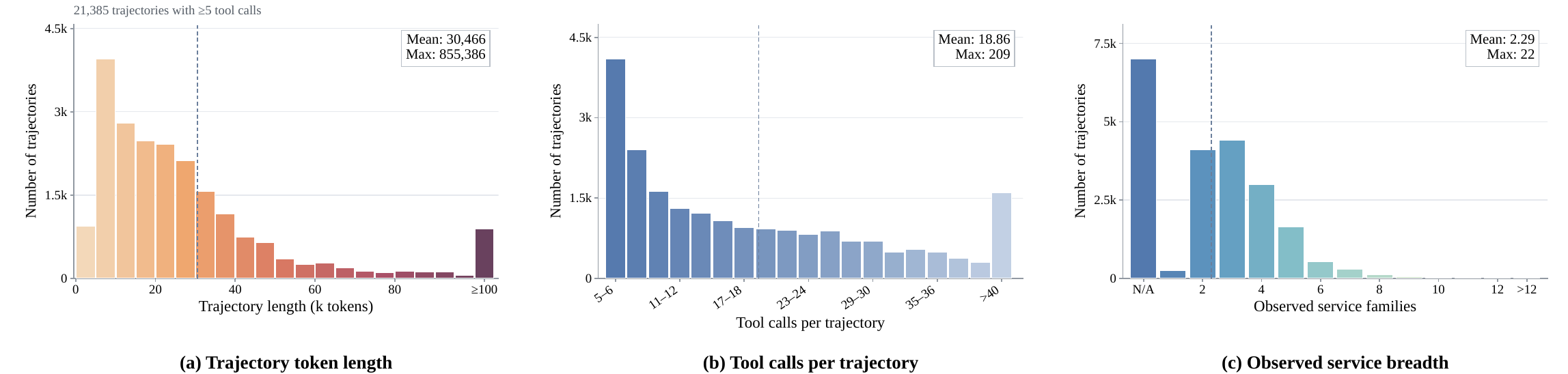}
\Description{Three histograms for 21,385 trajectories containing at least five tool calls: complete trajectory token length, individual tool-call count, and observed environment breadth.}

\captionsetup{font=jitcaptionsmall}
\caption{\textbf{Synthesized task and trajectory statistics.}
Distributions across 21,385 trajectories with at least five tool calls:
(a) trajectory token length,
(b) tool calls per trajectory, and
(c) environments per trajectory.}
\label{fig:toolspace-trajectory-statistics}
\end{figure}

To more comprehensively showcase ToolSpace,
Figures~\ref{fig:toolspace-inventory-statistics}
and~\ref{fig:toolspace-trajectory-statistics} characterize the static
statistics of ToolSpace's graph, environments, and synthesized tasks.

\textbf{(A) Graph Scale and Dependency Structure.}
Reconstructed from collected tool-call trajectories,
$\mathcal{G}_{\mathrm{TS}}$ contains 77{,}760 tool-call-pattern nodes derived from 2,042 distinct canonical tools. Its mean and maximum outdegrees (outgoing dependency edges per node) are 2.3 and 56, respectively.
Panel~(a) presents a word cloud of the environment types covered by
ToolSpace synthesis, including Gmail, Google Sheets, Slack, Google
Drive, Salesforce, and Airtable. Panel~(d) shows that data, state,
and retry edges account for 69\%, 25\%, and 6\%, respectively. The
dominance of data edges indicates that tool composition is driven primarily
by intermediate-result transfer, while state and retry edges capture state
prerequisites and failure recovery. Every edge is witnessed by adjacent
calls in real trajectories, providing execution-grounded tool reachability.

\textbf{(B) Environment Diversity and Selective Grounding.}
The reference inventory covers 804 real MCP environments.
Figure~\ref{fig:toolspace-inventory-statistics}(b) identifies
Communication \& General Utilities, Finance \& Markets, and Document \&
Design as the largest domains, accounting for 27.4\%, 11.2\%, and 10.1\%
of environments, respectively. Panel~(c) reports an average of 10.86
matched tools per environment and a maximum of 436; tools such as web
search, file reading, and email sending are available across multiple
environments. Panel~(e) distinguishes 453 operational tools (22.2\%)
from 1{,}589 non-operational tools (77.8\%), with structured record
mutation being the largest operational category at 95 tools. Among the
804 environments with assigned tools, Panel~(f) identifies 181 operational
(22.5\%) and 623 non-operational environments (77.5\%), with document
persistence being the largest operational category at 42 environments.
An environment is operational if it contains at least one tool capable of
modifying its state. These statistics further show that stateless environments dominate
the real-world environments synthesized by ToolSpace. ToolSpace
materializes databases on demand only for operational workspaces,
restores snapshots for state-query tools, and reuses cached returns
for stateless environments and tools, thereby reducing
database-synthesis requirements and costs relative to data-centric
methods.

\textbf{(C) Task Scale and Complexity.}
In total, ToolSpace synthesizes 34{,}376 trajectories.
Figure~\ref{fig:toolspace-trajectory-statistics} summarizes the 21{,}385
trajectories with at least five tool calls. Panel~(a) reports mean and
maximum lengths of 30{,}466 and 855{,}386 tokens, while Panel~(b) reports
an average of 18.86 tool calls per trajectory and a maximum of 209,
demonstrating that ToolSpace can synthesize long-horizon agent
execution trajectories. Panel~(c) shows an average
of 2.29 environments per trajectory and a maximum of 22, with at least
66.2\% of trajectories involving two or more environments. During synthesis,
tool chains are sampled through state-aware random walks, executed
step by step, and retained only after replay verification. The resulting
tasks therefore require both long-horizon tool orchestration and
consistent environment state.

\section{AutomationBench Pro and \texorpdfstring{$\tau^3$}{Tau-3}-Bench Banking Pro}
\label{sec:pro}
\begin{wrapfigure}{r}{0.47\textwidth}
\centering
\includegraphics[width=\linewidth]{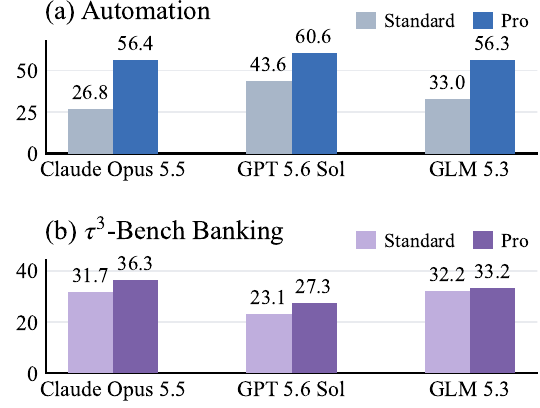}
\Description{Paired bars of average tool calls on AutomationBench versus AutomationBench Pro and on tau-3-Bench Banking versus tau-3-Bench Banking Pro for Claude Opus 5.5, GPT, and GLM 5.3.}
\caption{Average tool calls of frontier models on (a)~AutomationBench and (b)~$\tau^3$-Bench Banking (Standard vs.\ Pro).}
\label{fig:pro-complexity}
\end{wrapfigure}

To test whether ToolSpace can raise the cross\hyp{}environment and long\hyp{}horizon difficulty of frontier benchmarks, we anchor the pipeline to AutomationBench~\citep{automationbench} and $\tau^3$-Bench~Banking~\citep{tauknowledge}, showing the pipeline's flexible difficulty expansion.

\textbf{Benchmark Construction.} For a benchmark $b$ with original environment $e_b^{\mathrm{orig}}=(\mathcal{D}_b^{\mathrm{orig}},\mathcal{F}_b^{\mathrm{orig}})$, we collect its real tool calls $\mathcal{M}_b$, build a tool graph $\mathcal{G}_b$ by the rules of \S\ref{sec:graph}, and thereby link in tools outside the original benchmark that share data or state relations with benchmark tools, expanding a single environment into a cross\hyp{}environment cluster. We then restore or synthesize the necessary databases and snapshots (\S\ref{sec:space}), walk with larger $L$ and $K$ than the original benchmark, and reverse-generate tasks and verification scripts. After replay and deduplication we obtain AutomationBench Pro (300 tasks) and $\tau^3$-Bench~Banking Pro (98 tasks). To ensure that Pro benchmarks reflect genuine difficulty rather than task quality, three doctoral-level experts independently reviewed all candidate tasks and their verification scripts over a two-day period. Only tasks approved by at least two experts were retained. Inter-rater agreement was strong, with Fleiss' $\kappa = 0.82$ and an observed agreement of $93.1\%$.

\textbf{Data Statistics.} Figure~\ref{fig:pro-complexity} shows that Pro tasks require substantially more tool calls than the standard splits on both benchmarks. On AutomationBench, average tool calls rise sharply for every model, from 26.8--43.6 under Standard to 56.3--60.6 under Pro. On $\tau^3$-Bench~Banking, average calls likewise rise for every model, from 23.1--32.2 to 27.3--36.3, fully confirming that the added difficulty arises from longer and more frequent tool orchestration.

\FloatBarrier

\section{Experiments}
\label{sec:exp}

\subsection{Experimental Setup}
\label{sec:setup}

\textbf{Benchmarks.} We evaluate on seven long\hyp{}horizon tool\hyp{}use benchmarks, referred to by their full names throughout: \textbf{AutomationBench}~\citep{automationbench} (workflow automation), \textbf{SkillsBench}~\citep{skillsbench} (skill composition), \textbf{JobBench}~\citep{jobbench} (job-style professional tasks), \textbf{$\tau^3$-Bench Banking}~\citep{tauknowledge} (banking tool--dialogue interaction), the \textbf{Berkeley Function Calling Leaderboard V4 (BFCL V4)}~\citep{bfcl} (function calling), \textbf{Harvey LAB}~\citep{harveylab} (legal agent workflows), and \textbf{OfficeQA Pro}~\citep{officeqa} (office document question answering).

\textbf{Baselines.}
We evaluate four model groups:
(i)~\emph{proprietary frontier models}: Claude Opus~5~\citep{claude}
and GPT~5.6 Sol~\citep{openai};
(ii)~\emph{open-weight models above 200B parameters}: GLM~5.3~\citep{glm53},
GLM~5.2~\citep{glm52}, DeepSeek-V4-Pro-0813~\citep{deepseek},
Kimi~K3~\citep{kimi}, and Qwen~3.8 Max~\citep{qwen38max};
(iii)~\emph{7--8B environment-scaling models}: EnvScaler-8B~\citep{envscaler},
AWM-8B~\citep{awm}, Simulator-8B~\citep{simia},
Agent-World-8B~\citep{dong2026agent}, Agent-Reflex-8B~\citep{agentreflex},
and their Qwen3-8B backbone~\citep{qwen3}; and
(iv)~\emph{27--35B open-weight models}: Qwen3.6-27B~\citep{qwen3.6-27b},
Qwen3.6-35B-A3B~\citep{qwen36}, Nex-N2 Mini~\citep{nex},
Muse-Glimmer-30B~\citep{museglimmer}, and
WorkRover-35B-A3B~\citep{workrover}.
For the Pro benchmark, we further evaluate Claude Opus~5.5~
\citep{anthropic2026claudeopus55}, GPT~6.1 Sol~\citep{openai2026gpt61sol},
and DeepSeek-V4.1-Flash~\citep{deepseekai2026deepseekv41flash}.

\textbf{Implementation Details.}
\label{sec:impl-main}
We train long-horizon agents on three backbones, Qwen3-8B, Qwen3.6-35B-A3B, and GLM-5.2-744B-A40B (the resulting 744B model is denoted ToolSpace-744B-A40B). Detailed ToolSpace statistics are reported in \S\ref{app:toolspace-statistics}.

\textbf{Trajectory collection and environment synthesis.}
We collect over 200{,}000 agent--tool interaction trajectories as raw material for ToolSpace construction, with strict quality control during collection. We perform decontamination at both the task level and the trajectory level: any collected trajectory whose task, tool-call sequence, or initial state overlaps with an evaluation test set is excluded, ensuring that the synthesized data are free of test-set contamination. All data sources and MCP services used are covered by open-source licenses. We mine executable environments following the agentic exploration procedure and settings of Agent-World~\cite{dong2026agent}.

\textbf{Task and trajectory synthesis.}
We use Kimi~K3 (K3) for both teacher-trajectory generation and reverse verifiable task synthesis. All synthesized samples are validated against $V$ using verification scripts, yielding 34{,}376 SFT samples and 653 RL samples from the same ToolSpace synthesis pipeline, except that RL candidates undergo additional Pass@16 screening with the 35B model, retaining only candidates with one or two correct rollouts to ensure informative RL training.

\textbf{Training and evaluation.}
All data-synthesis baselines use the same SFT configuration: the 8B and 35B models are trained for one epoch on 16 NVIDIA GPUs, while the 744B model is trained for one epoch on 256 GPUs. For agent RL, we adopt the GRPO objective~\cite{shao2024deepseekmath},
\begin{equation}
\label{eq:grpo}
\begin{aligned}
J_{\mathrm{GRPO}}(\theta)=\mathbb{E}_{x\sim\mathcal{D},\{y_i\}_{i=1}^G\sim\pi_{\theta_{\mathrm{old}}}(\cdot\mid x)}\Bigg[\frac{1}{G}\sum_{i=1}^{G}\frac{1}{|y_i|}\sum_{t=1}^{|y_i|}
&\min\big(r_{i,t}(\theta)\hat{A}_i,\ \mathrm{clip}(r_{i,t}(\theta),c_{\mathrm{low}},c_{\mathrm{high}})\hat{A}_i\big)\Bigg],
\end{aligned}
\end{equation}
where $r_{i,t}(\theta)$ is the token-level importance ratio between the current and rollout policies, and $c_{\mathrm{low}}$ and $c_{\mathrm{high}}$ are the clipping bounds, set to 0.8 and 1.28 in our training. The advantage and the reward are defined as
\begin{equation}
\label{eq:advantage-reward}
\hat{A}_i=\frac{R_i-\mu_g}{s_g+10^{-6}},\qquad
R_i=\mathbb{I}\big[\mathrm{Execute}(V_{\mathrm{code}}(y_i,y_i^*))\big]\in\{0,1\},
\end{equation}
where $\mu_g$ and $s_g$ are the mean and standard deviation of $\{R_i\}_{i=1}^G$ within rollout group $g$, i.e., the $G$ rollouts sampled for the same prompt, and $R_i$ is 1 only when execution-based verification of the completed trajectory passes, and 0 otherwise. Training runs for 48 steps with a learning rate of $2\times10^{-6}$, a total training batch size of 256, and a rollout group size of 16, in a fully asynchronous (full-async) manner with a maximum policy staleness of 4. Inference uses the official recommended settings, each benchmark follows its standard evaluation protocol, and all ToolSpace results are averaged over three independent runs.

\textbf{Evaluation settings and result provenance.}
We prioritize official leaderboard results, followed by provider-reported scores under the official protocol and, when unavailable for open-weight models, local reproduction using the official harness, split, metric, tool-call protocol, and recommended inference settings. Local results are averaged over three independent runs; external scores are used as published.

\subsection{Main Results}
\begin{table}[t]
\centering
\caption{Performance on seven tool-use benchmarks. Bold/underline mark the best/second-best score per column \emph{within each same-scale pool} (the 8B, 27--35B, and 744B pools, each including the ToolSpace model(s) at that scale). ``--'' indicates results not officially reported.}

\label{tab:main}
\begingroup
\definecolor{ToolSpaceRow}{RGB}{242,237,253}
\definecolor{ToolSpaceRLRow}{RGB}{229,219,249}
\setlength{\tabcolsep}{2.3pt}
\renewcommand{\arraystretch}{0.98}
\footnotesize
\begin{tabularx}{\textwidth}{l>{\centering\arraybackslash}c >{\centering\arraybackslash}c *{6}{>{\centering\arraybackslash}X}>{\centering\arraybackslash}c}
\toprule
\textbf{Model} & \textbf{Size} & \shortstack{\textbf{Automation}\\\textbf{Bench}} & \shortstack{\textbf{Skills}\\\textbf{Bench}} & \shortstack{\textbf{Job}\\\textbf{Bench}} & \shortstack{$\symbf{\tau}^{3}$\textbf{-Bench}\\\textbf{Banking}} & \textbf{BFCL V4} & \shortstack{\textbf{Harvey}\\\textbf{LAB}} & \shortstack{\textbf{OfficeQA}\\\textbf{Pro}} & \textbf{Avg.} \\
\midrule
\multicolumn{10}{@{}l}{\textbf{\textit{Proprietary Models}}} \\
\mlogo{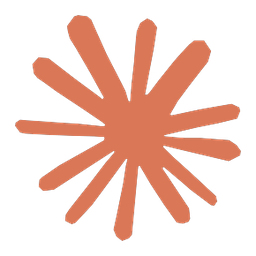}Claude Opus 5 & -- & 45.6 & 63.7 & \textbf{68.0} & \underline{48.7} & \textbf{80.6} & 22.7 & \textbf{67.2} & \textbf{56.6} \\
\mlogo{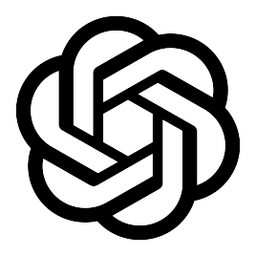}GPT 5.6 Sol & -- & 47.1 & 62.5 & 45.4 & 46.9 & 68.8 & 5.0 & 66.4 & 48.9 \\
\midrule
\multicolumn{10}{@{}l}{\textbf{\textit{Open-Weight Models ($>$200B total parameters)}}} \\
\mlogo{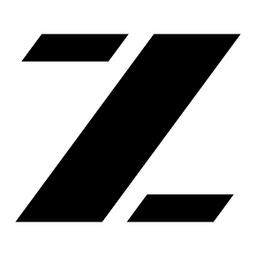}GLM 5.3 & 744B & \underline{52.5} & 63.3 & \underline{58.2} & 40.2 & 74.1 & 22.5 & 65.8 & 53.8 \\
\mlogo{glm}GLM 5.2 & 744B & 29.3 & 56.7 & 43.4 & 37.1 & 75.6 & 13.3 & 56.9 & 44.6 \\
\mlogo{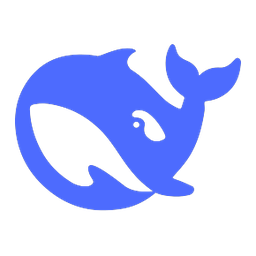}DeepSeek-V4-Pro-0813 & 1.7T & 41.7 & 65.0 & 54.1 & 44.3 & 71.4 & 6.7 & 66.2 & 49.9 \\
\mlogo{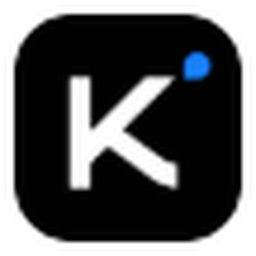}Kimi K3 & 2.8T & 44.8 & 51.9 & 54.3 & 37.1 & 69.1 & \underline{25.4} & 65.4 & 49.7 \\
\mlogo{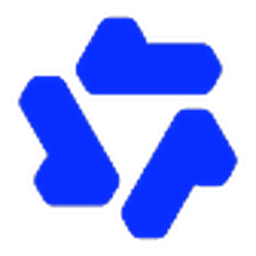}Qwen 3.8 Max & 2.4T & 49.7 & \underline{66.7} & 52.7 & \textbf{55.2} & 66.8 & 24.2 & 66.2 & \underline{54.5} \\
\midrule
\multicolumn{10}{@{}l}{\textbf{\textit{Environment Scaling Baselines (7--8B)}}} \\
\mlogo{qwen}Qwen3-8B & 8B & 0.0 & 0.1 & 0.1 & 4.1 & 42.8 & 0.0 & 3.0 & 7.2 \\
\mlogopdf{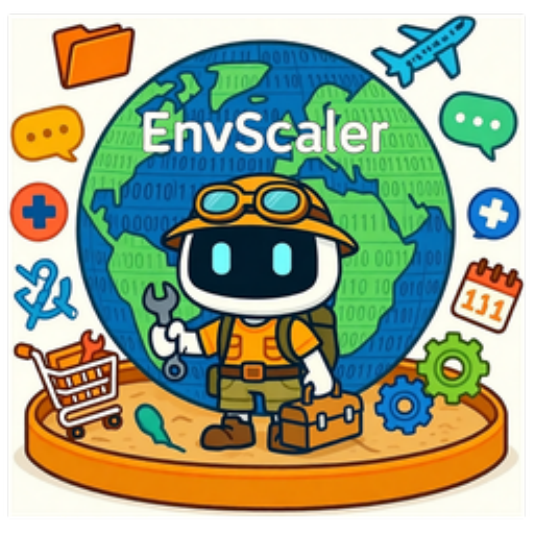}EnvScaler-8B & 8B & 0.0 & 1.6 & 1.1 & 7.2 & 47.5 & 0.0 & 1.5 & 8.4 \\
\mlogo{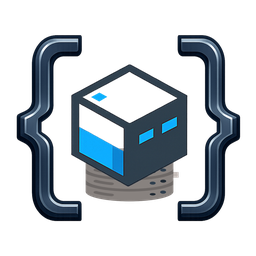}AWM-8B & 8B & 0.0 & 1.4 & 0.5 & 5.2 & 42.0 & 0.0 & 5.3 & 7.8 \\
\mlogo{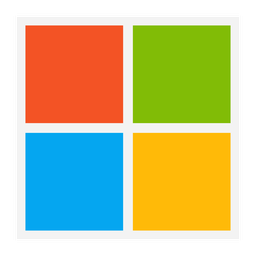}Simulator-8B & 8B & 0.0 & 0.4 & 0.1 & 1.0 & 20.5 & 0.0 & \underline{6.0} & 4.0 \\
\mlogo{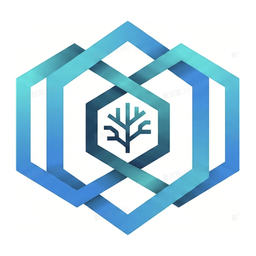}Agent-World-8B & 8B & 0.3 & 2.5 & 0.7 & 6.2 & 49.0 & 0.0 & 3.8 & 8.9 \\
\mlogo{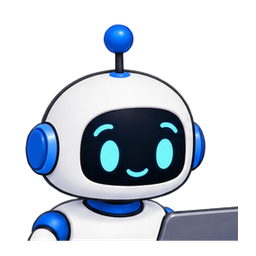}Agent-Reflex-8B & 8B & 0.5 & 3.3 & 0.8 & 3.1 & 51.3 & 0.0 & 5.3 & 9.2 \\
\midrule
\multicolumn{10}{@{}l}{\textbf{\textit{Open-Weight Foundation Models (27--35B)}}} \\
\mlogo{qwen}Qwen3.6-27B & 27B & 10.7 & 22.1 & 5.6 & 15.8 & 58.5 & 0.8 & 38.4 & 21.7 \\
\mlogo{qwen}Qwen3.6-35B-A3B & 35B & 8.2 & 30.7 & 6.4 & 9.3 & 60.0 & 0.8 & 45.9 & 23.0 \\
\mlogo{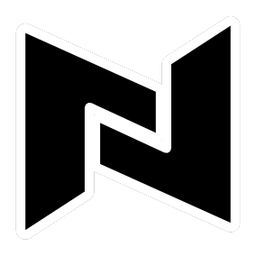}Nex-N2 Mini & 35B & 13.2 & 16.8 & 11.5 & 18.6 & 63.3 & 0.0 & 38.4 & 23.1 \\
\mlogo{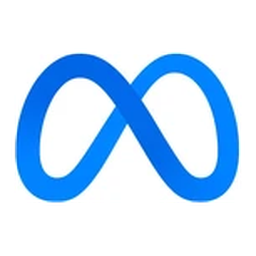}Muse-Glimmer-30B & 30B & 14.2 & 32.9 & 3.5 & 11.3 & 55.5 & 0.0 & 13.5 & 18.7 \\
\mlogo{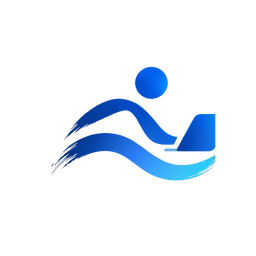}WorkRover-35B-A3B & 35B & \underline{28.7} & -- & \underline{36.9} & -- & -- & -- & \underline{48.9} & -- \\
\midrule
\multicolumn{10}{@{}l}{\textbf{\textit{ToolSpace (Ours)}}} \\
\rowcolor{ToolSpaceRow}
\tslogo ToolSpace-8B & 8B & \underline{32.3} & \underline{3.9} & \underline{13.1} & \underline{10.3} & \underline{65.3} & 0.0 & \textbf{12.0} & \underline{19.6} \\
\rowcolor{ToolSpaceRLRow}
\quad\textit{+ Agent RL} & 8B & \textbf{34.3} & \textbf{4.1} & \textbf{15.5} & \textbf{11.3} & \textbf{67.5} & 0.0 & \textbf{12.0} & \textbf{20.7} \\
\rowcolor{ToolSpaceRow}
\tslogo ToolSpace-35B-A3B & 35B & \textbf{45.8} & \underline{42.4} & \textbf{40.0} & \underline{34.0} & \underline{74.3} & \underline{1.7} & \underline{48.9} & \underline{41.0} \\
\rowcolor{ToolSpaceRLRow}
\quad\textit{+ Agent RL} & 35B & \textbf{45.8} & \textbf{45.3} & \textbf{40.0} & \textbf{42.3} & \textbf{74.8} & \textbf{2.5} & \textbf{51.9} & \textbf{43.2} \\
\rowcolor{ToolSpaceRow}
\tslogo ToolSpace-744B-A40B & 744B & \textbf{54.7} & \textbf{67.8} & 57.5 & 43.3 & \underline{78.4} & \textbf{27.5} & \underline{67.1} & \textbf{56.6} \\
\bottomrule
\end{tabularx}
\endgroup\end{table}

\textbf{Long-horizon tool use and office productivity.}
\label{sec:main-results}
Table~\ref{tab:main} reports performance on seven benchmarks. ToolSpace is strong across model scales; three observations summarize this.

\textbf{(1) Environment scaling learns the form of function calling, not tool orchestration.} All 7--8B environment scaling baselines remain weak ($<$10) on the cross\hyp{}environment, long\hyp{}horizon suites AutomationBench, $\tau^3$-Bench~Banking, JobBench, and Harvey LAB. Single-environment data therefore correct the \emph{form} of calls without teaching cross\hyp{}environment composition.

\textbf{(2) ToolSpace exhibits cross\hyp{}environment long\hyp{}horizon execution across scales.} Gains hold in each domain: ToolSpace-8B leads same-size environment scaling baselines on real productivity suites such as AutomationBench, JobBench, and OfficeQA Pro; ToolSpace-35B-A3B exceeds its Qwen backbone on all seven benchmarks, with AutomationBench rising by more than 35 points (8.2$\rightarrow$45.8). The pattern persists to 744B: ToolSpace-744B-A40B surpasses GLM~5.3, GPT~5.6 Sol, and its teacher policy K3 on average, and matches Claude Opus~5. These consistent gains across model scales, including performance beyond the teacher, argue against simple capability distillation and instead indicate that ToolSpace teaches transferable tool-composition structure for cross\hyp{}environment execution.

\textbf{(3) On-policy agent RL further unlocks capacity.} SFT alone lifts ToolSpace by 18.0 points over the 35B backbone (23.0$\rightarrow$41.0). On-policy RL on ToolSpace data adds a further 2.2 points at 35B (41.0$\rightarrow$43.2), with the largest gain on $\tau^3$-Bench~Banking (34.0$\rightarrow$42.3), showing that environment RL still leaves room for improvement. This indicates that on-policy environment RL can, through exploration, further unlock the multi-tool orchestration potential of SFT-trained models.
\subsection{Results on AutomationBench Pro and $\tau^3$-Bench Banking Pro}
\definecolor{DeltaCol}{RGB}{253,232,232}
\begin{table}[t]
\caption{\textbf{Results on AutomationBench Pro and $\tau^3$-Bench Banking Pro.} Std.\ denotes the standard split; Pro denotes our upgraded split. $\Delta$ is Pro$-$Standard.
Std.\ scores are taken from the Atria Dawn public leaderboard (\url{https://atria-asi.ai/\#benchmark-results}).
The scores of DeepSeek-V4.1-Flash follow their official release notes.}
\label{tab:pro}
\centering
\scriptsize
\renewcommand{\arraystretch}{1.18}
\setlength{\tabcolsep}{2.2pt}
\begin{tabularx}{\textwidth}{@{}l >{\centering\arraybackslash}X >{\centering\arraybackslash}X >{\columncolor{DeltaCol}\centering\arraybackslash}X >{\centering\arraybackslash}X >{\centering\arraybackslash}X >{\columncolor{DeltaCol}\centering\arraybackslash}X@{}}
\toprule
\multirow{2}{*}{\textbf{Frontier model}} & \multicolumn{3}{c}{\textbf{AutomationBench}} & \multicolumn{3}{c}{\textbf{$\tau^3$-Bench~Banking}} \\
\cmidrule(lr){2-4}\cmidrule(l){5-7}
& \textbf{Std.} & \textbf{Pro} & \textbf{$\Delta$$\downarrow$} & \textbf{Std.} & \textbf{Pro} & \textbf{$\Delta$$\downarrow$} \\
\midrule
\mlogo{claude}\textbf{Claude Opus 5.5} & 59.2 & 26.3 & $-32.9$ & 51.5 & 35.7 & $-15.8$ \\
\mlogo{openai}\textbf{GPT 6.1 Sol} & 50.8 & 24.7 & $-26.1$ & 46.4 & 29.6 & $-16.8$ \\
\mlogo{deepseek}\textbf{DeepSeek-V4.1-Flash-1017} & 60.5 & 22.3 & $-38.2$ & 51.3 & 36.9 & $-14.4$ \\
\mlogo{deepseek}\textbf{DeepSeek-V4.1-Flash} & 54.8 & 20.0 & $-34.8$ & 50.3 & 35.7 & $-14.6$ \\
\mlogo{glm}\textbf{GLM 5.3} & 52.5 & 22.7 & $-29.8$ & 40.2 & 20.4 & $-19.8$ \\
\mlogo{qwen}\textbf{Qwen 3.8 Max} & 49.7 & 24.0 & $-25.7$ & 55.2 & 29.6 & $-25.6$ \\
\mlogo{glm}\textbf{GLM 5.2} & 29.3 & 18.0 & $-11.3$ & 37.1 & 12.2 & $-24.9$ \\
\bottomrule
\end{tabularx}
\end{table}

\label{sec:exp-pro}
To test whether ToolSpace raises frontier-benchmark difficulty, Table~\ref{tab:pro} compares seven frontier models on the two Pro splits with Standard. All seven frontier models drop substantially from Standard to Pro. Even Claude Opus~5.5 and GPT~6.1 Sol fall by 32.9 and 26.1 points on AutomationBench, and by 15.8 and 16.8 points on $\tau^3$-Bench~Banking, respectively; the largest AutomationBench drop reaches 38.2 points. Together with Figure~\ref{fig:pro-complexity}, where Pro tasks require more tool calls, this shows that, as cross\hyp{}environment and tool-execution difficulty rises sharply, even frontier models still struggle on these tasks, further underscoring ToolSpace's flexible difficulty control.

\subsection{Quantitative Analyses}
\label{sec:quant}

\begin{figure}[t]
\centering
\includegraphics[width=\textwidth]{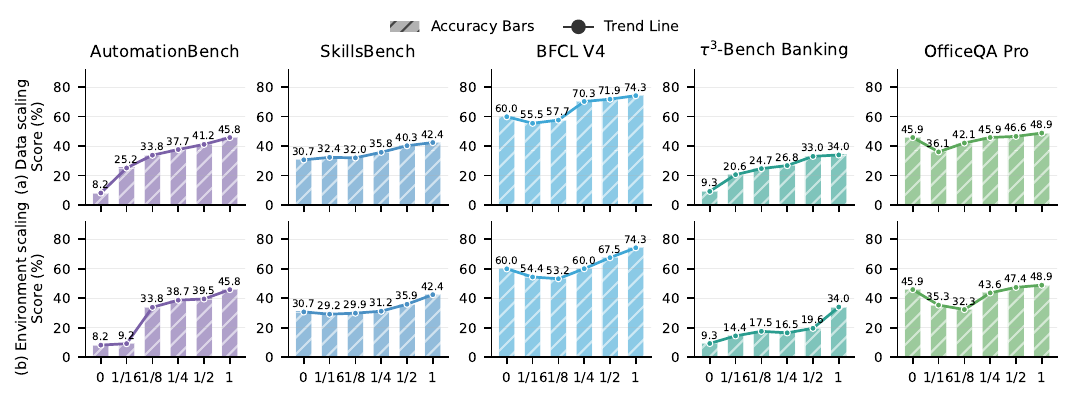}
\Description{Two rows of five benchmark panels. Top row (a) shows
performance as the training-data fraction increases. Bottom row (b)
shows performance as the training-environment fraction increases.}

\caption{\textbf{Data and environment scaling results.}
(a) Data scaling by training-data fraction.
(b) Environment scaling by training-environment fraction.
Zero denotes the untuned Qwen3.6-35B-A3B backbone.}
\label{fig:scaling}
\end{figure}

\textbf{Environment Scaling Analysis.} To examine whether the breadth of the tool space matters independently of data volume, we fine-tune Qwen3.6-35B-A3B on data synthesized from increasing fractions of ToolSpace environments, from $1/16$ to the full environment set. The backbone, training configuration, and evaluation protocol are held fixed, and zero denotes the untuned backbone. Figure~\ref{fig:scaling}(b) reports the results. Despite fluctuations at intermediate fractions, performance improves overall with environment scale and reaches the full-setting optimum, following a clear scaling trend. The increase on AutomationBench between $1/16$ and $1/8$ is particularly sharp, from 9.2 to 33.8. This suggests that new environment clusters contribute new cross\hyp{}environment edges, and that space breadth is the limiting factor when coverage is low. In contrast, $\tau^3$-Bench~Banking improves more slowly under environment scaling, which is consistent with its reliance on a narrower set of stateful banking tools. This supports the central motivation of ToolSpace: performance scales with the breadth of environments that witnessed chains connect. From the perspective of sustained scaling returns, AutomationBench gains about $+4$ points per doubling beyond $1/8$ on average, despite a plateau between $1/4$ and $1/2$, while $\tau^3$-Bench~Banking shows its largest jump only at the full environment set (19.6$\to$34.0), indicating that highly connected environment clusters contribute late.

\begin{figure}[t]
\centering
\includegraphics[width=\linewidth]{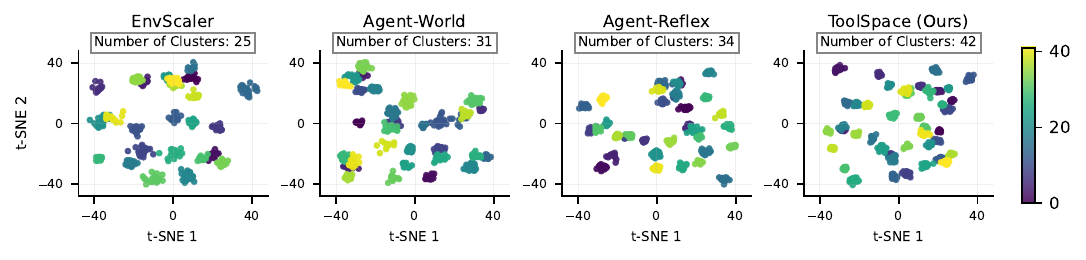}
\Description{Four panels show question embeddings for EnvScaler, Agent-World, Agent-Reflex, and ToolSpace, colored by DBSCAN cluster, each annotated with its number of clusters.}

\caption{\textbf{Task-Synthesis Diversity Visualization.} Synthesized task embeddings colored by DBSCAN cluster.}
\label{fig:diversity}
\end{figure}

\textbf{Data Scaling Analysis.}
To examine whether the benefit of ToolSpace grows with the amount of synthesized data, we fine-tune Qwen3.6-35B-A3B on increasing fractions of the ToolSpace SFT dataset, namely $1/16$, $1/8$, $1/4$, $1/2$, and the full dataset. The backbone, training configuration, and evaluation protocol are held fixed, and zero denotes the untuned backbone. Figure~\ref{fig:scaling}(a) reports the results. Performance improves overall as more data are used, and the full dataset performs best on all five benchmarks. SkillsBench shows a small intermediate decrease from 32.4 at $1/16$ to 32.0 at $1/8$. AutomationBench rises from 25.2 at $1/16$ to 45.8 with the full dataset, and $\tau^3$-Bench~Banking rises from 20.6 to 34.0. The gain is therefore not explained by a small set of high-quality examples. Each additional fraction contributes new witnessed tool chains and dependency structures, which is consistent with our motivation that capability grows from the composition space covered by the data. The small decreases below the backbone at the $1/16$ fraction on BFCL V4 and OfficeQA Pro indicate interference when the SFT mixture is too small, and the trend becomes stable once the data fraction increases. This demonstrates the scalability of ToolSpace's synthesized data. From the perspective of sustained scaling returns, the step from $1/2$ to the full dataset still improves all five benchmarks (AutomationBench $+4.6$), with no plateau, indicating that the current synthesis scale has not yet hit the returns ceiling of data volume.

\begin{figure}[t]
\centering
\begin{minipage}{0.49\textwidth}
\centering
\includegraphics[width=\linewidth]{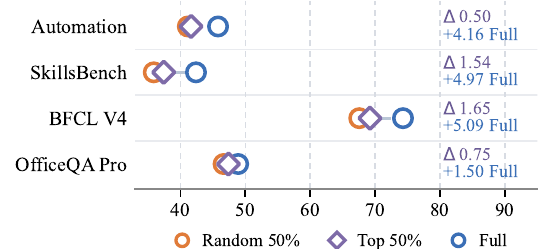}
\end{minipage}\hfill
\begin{minipage}{0.49\textwidth}
\centering
\includegraphics[width=\linewidth]{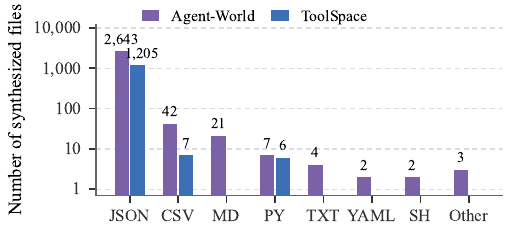}
\end{minipage}
\caption{\textbf{Environment relevance and synthesis cost.}
(a) Effect of environment relevance: Full, Top 50\%, and Random 50\% environments; $\Delta$ is the Top--Random gap.
(b) Environment synthesis cost proxy: synthesized files by type over 500 sampled environments.}
\label{fig:envsel-cost}
\end{figure}

\textbf{Environment Relevance Analysis.} To verify that ToolSpace's gains do not stem from overfitting to benchmark-like environments, we rank environment clusters by relevance to each benchmark using BGE~\citep{bge} embeddings, and compare SFT on (i) the full data, (ii) only the top-50\% most relevant environments (Top-50\%), and (iii) a random 50\% of environments (Random-50\%) (Figure~\ref{fig:envsel-cost}(a)). If the gains were driven by benchmark-specific environment fitting, Top-50\% should substantially outperform Random-50\%. Random-50\% uses the same number of environments but excludes many of the most relevant ones. Instead, the two differ by at most 1.7 points on AutomationBench, SkillsBench, BFCL V4, and OfficeQA Pro, and both approach full-data performance using only half the environments. ToolSpace's gains therefore do not hinge on benchmark-aligned environments, but reflect transferable tool-composition structure.

\textbf{Task-Synthesis Diversity Visualization.}\label{sec:diversity}
We examine whether ToolSpace spans a broader space of environments and tasks than prior environment-scaling methods by randomly sampling 2k questions from four SFT corpora (EnvScaler, Agent-World, Agent-Reflex, and ToolSpace). We employ BGE-M3~\citep{bge} as the embedding model, apply t-SNE for dimensionality reduction, and use DBSCAN~\citep{dbscan} for clustering. As shown in Figure~\ref{fig:diversity}, ToolSpace forms the largest number of question clusters (42), compared with EnvScaler, Agent-World, and Agent-Reflex (25, 31, and 34, respectively), while retaining compact local groupings and broader coverage of the embedding space. This pattern suggests that ToolSpace expands task-space coverage beyond prior environment-scaling methods. We attribute the gain to edge-weighted sampling over observed tool compositions, which connects a wider range of environments.

\textbf{Environment Synthesis Cost Analysis.}\label{sec:cost_analysis} We estimate ToolSpace's environment synthesis cost savings using the number of synthesized files as a proxy. From 500 environments randomly sampled from Agent-World's public library, we identify the files that ToolSpace would need to synthesize. Because file generation and validation complexity may vary, this proxy does not directly measure token, wall-clock, API, or compute cost.

As shown in Figure~\ref{fig:envsel-cost}(b), Agent-World synthesizes all 2{,}724
files in the sample, whereas ToolSpace synthesizes only 1{,}218 files
associated with operation-bearing environments---1{,}506 fewer files,
or a 55.3\% reduction. Specifically, JSON files decrease from 2{,}643
to 1{,}205 and CSV files from 42 to 7.

This reduction follows from ToolSpace's selective-grounding strategy:
persistent artifacts are materialized only for operation-bearing
environments, while recorded returns are reused for stateless interactions.
The lower artifact-generation workload indicates reduced environment
synthesis cost, although exact token, time, and compute savings are not
measured.

\begin{table}[t]
\caption{\textbf{Component ablation.} Darker shading marks a larger within-column drop from Full.}

\label{tab:ablation}
\centering
\scriptsize
\renewcommand{\arraystretch}{1.20}
\setlength{\tabcolsep}{2pt}
\begin{tabularx}{\textwidth}{@{}>{\raggedright\arraybackslash}p{0.22\textwidth}>{\hsize=1.2\hsize\linewidth=\hsize\centering\arraybackslash}X*{4}{>{\hsize=.95\hsize\linewidth=\hsize\centering\arraybackslash}X}@{}}
\toprule
\textbf{Variant} & {\scriptsize\shortstack{\textbf{Automation}\\\textbf{Bench}}} & {\scriptsize\shortstack{\textbf{Skills}\\\textbf{Bench}}} & {\scriptsize\shortstack{\textbf{BFCL}\\\textbf{V4}}} & {\scriptsize\shortstack{\textbf{$\tau^3$-Bench}\\\textbf{Banking}}} & {\scriptsize\shortstack{\textbf{OfficeQA}\\\textbf{Pro}}} \\
\midrule
\textbf{Full ToolSpace} & \cellcolor[RGB]{242,237,253}\textbf{45.8} & \cellcolor[RGB]{242,237,253}\textbf{42.4} & \cellcolor[RGB]{242,237,253}\textbf{74.3} & \cellcolor[RGB]{242,237,253}\textbf{34.0} & \cellcolor[RGB]{242,237,253}\textbf{48.9} \\
w/o state edges & \cellcolor[RGB]{126,107,168}\textcolor{white}{36.5} & \cellcolor[RGB]{155,140,189}\textcolor{white}{38.0} & \cellcolor[RGB]{198,189,219}\textcolor{black}{70.1} & \cellcolor[RGB]{126,107,168}\textcolor{white}{21.5} & \cellcolor[RGB]{172,159,200}\textcolor{white}{44.0} \\
w/o data edges & \cellcolor[RGB]{146,130,182}\textcolor{white}{38.2} & \cellcolor[RGB]{146,130,182}\textcolor{white}{37.5} & \cellcolor[RGB]{126,107,168}\textcolor{white}{62.4} & \cellcolor[RGB]{200,191,220}\textcolor{black}{29.8} & \cellcolor[RGB]{126,107,168}\textcolor{white}{40.6} \\
w/o retry edges & \cellcolor[RGB]{214,207,231}\textcolor{black}{43.9} & \cellcolor[RGB]{214,207,231}\textcolor{black}{41.2} & \cellcolor[RGB]{223,217,237}\textcolor{black}{72.8} & \cellcolor[RGB]{224,219,238}\textcolor{black}{32.6} & \cellcolor[RGB]{224,218,238}\textcolor{black}{47.9} \\
w/o rollback & \cellcolor[RGB]{180,168,206}\textcolor{black}{41.0} & \cellcolor[RGB]{189,178,213}\textcolor{black}{39.8} & \cellcolor[RGB]{211,204,229}\textcolor{black}{71.5} & \cellcolor[RGB]{202,194,222}\textcolor{black}{30.1} & \cellcolor[RGB]{203,194,222}\textcolor{black}{46.3} \\
Uniform walk (no $w$) & \cellcolor[RGB]{163,149,194}\textcolor{white}{39.6} & \cellcolor[RGB]{172,159,201}\textcolor{white}{38.9} & \cellcolor[RGB]{195,186,217}\textcolor{black}{69.8} & \cellcolor[RGB]{187,176,212}\textcolor{black}{28.4} & \cellcolor[RGB]{186,176,211}\textcolor{black}{45.1} \\
LLM-judged edges & \cellcolor[RGB]{141,125,179}\textcolor{white}{37.8} & \cellcolor[RGB]{126,107,168}\textcolor{white}{36.4} & \cellcolor[RGB]{168,155,198}\textcolor{white}{66.9} & \cellcolor[RGB]{172,159,201}\textcolor{white}{26.7} & \cellcolor[RGB]{165,151,196}\textcolor{white}{43.5} \\
\bottomrule
\end{tabularx}
\end{table}
\textbf{Component Ablation.} Beyond which environments are selected and how much they cost to synthesize, we finally examine which ToolSpace components drive the gains. Table~\ref{tab:ablation} reports a component ablation of ToolSpace at 35B, where w/o denotes removing or replacing one specific component. We make three observations. \textbf{(1)} Removing any component degrades all five benchmarks, showing that the components are complementary rather than redundant. \textbf{(2)} The two structural edge types divide the labor: state edges matter most for $\tau^3$-Bench Banking and AutomationBench, where state preconditions decide whether a chain is executable at all, while data edges matter most for BFCL V4 and OfficeQA Pro, where argument provenance carries the task. Their comparable average drops (7.1 vs.\ 7.4 points) fall on disjoint benchmarks, confirming that state dependencies, not data flow alone, enable cross\hyp{}environment execution. \textbf{(3)} How edges are obtained and traversed matters as much as which edges exist: LLM-judged edges degrade every benchmark almost as much as removing state edges, so edges must be witnessed by execution rather than merely plausible; rollback and edge-weighted sampling contribute in between, and retry edges have the smallest effect, consistent with their role in robustness to noisy tools.

\section{Conclusion}
\label{sec:conclusion}
In this paper, we present ToolSpace, a tool-centric framework for cross-environment agent data synthesis. ToolSpace mines data, state, and retry dependencies between adjacent calls in execution traces from real MCP services and open agent corpora, using deterministic rules rather than LLM judgment, and clusters the resulting weighted tool-composition graph into environments. Databases are synthesized only for state-mutating environments, while query and stateless tools replay recorded returns. ToolSpace then performs state-aware random walks with rollback and decompiles each chain into a natural-language task with assertion-based checks, keeping only samples that replay successfully from the initial state. Applied to AutomationBench and $\tau^3$-Bench Banking, the pipeline produces harder Pro versions on which seven frontier models drop by up to 38.2 points. Across seven long-horizon tool-use benchmarks, ToolSpace raises Qwen3.6-35B-A3B from 23.0 to 41.0 average score (43.2 with RL) and GLM-5.2 from 44.6 to 56.6, on par with Claude Opus 5. Ablations show that execution-witnessed edges clearly outperform LLM-judged ones, and performance keeps improving with data and environment scale.

\FloatBarrier
\clearpage
\bibliographystyle{plainnat}
\bibliography{references}

\begin{thebibliography}{52}
\providecommand{\natexlab}[1]{#1}
\providecommand{\url}[1]{\texttt{#1}}
\expandafter\ifx\csname urlstyle\endcsname\relax
  \providecommand{\doi}[1]{doi: #1}\else
  \providecommand{\doi}{doi: \begingroup \urlstyle{rm}\Url}\fi

\bibitem[{Anthropic}(2024)]{mcp}
{Anthropic}.
\newblock Model context protocol ({MCP}) specification.
\newblock \url{https://modelcontextprotocol.io}, 2024.
\newblock Accessed 26 Oct. 2026.

\bibitem[{Anthropic}(2026{\natexlab{a}})]{anthropic2026claudeopus55}
{Anthropic}.
\newblock System card: Claude opus 5.5.
\newblock \url{https://www-cdn.anthropic.com/fc1b44717c85dc068bc6ba5024219938094694bd/Claude\%20Opus\%205.5\%20System\%20Card.pdf}, September 2026{\natexlab{a}}.
\newblock Released September 22, 2026.

\bibitem[{Anthropic}(2026{\natexlab{b}})]{claude}
{Anthropic}.
\newblock Introducing {Claude Opus 5}.
\newblock \url{https://www.anthropic.com/news/claude-opus-5}, 2026{\natexlab{b}}.
\newblock Official announcement, 24 Jul. 2026; accessed 26 Oct. 2026.

\bibitem[Chen et~al.(2024)Chen, Xiao, Zhang, Luo, Lian, and Liu]{bge}
Jianlyu Chen, Shitao Xiao, Peitian Zhang, Kun Luo, Defu Lian, and Zheng Liu.
\newblock {M}3-embedding: Multi-linguality, multi-functionality, multi-granularity text embeddings through self-knowledge distillation.
\newblock In Lun-Wei Ku, Andre Martins, and Vivek Srikumar, editors, \emph{Findings of the {A}ssociation for {C}omputational {L}inguistics: {ACL} 2024}, pages 2318--2335, Bangkok, Thailand, August 2024. Association for Computational Linguistics.
\newblock \doi{10.18653/v1/2024.findings-acl.137}.
\newblock \url{https://aclanthology.org/2024.findings-acl.137/}.

\bibitem[{DeepSeek-AI}(2026)]{deepseekai2026deepseekv41flash}
{DeepSeek-AI}.
\newblock Deepseek-v4.1-flash: Pushing the limits of kv cache compression, 2026.

\bibitem[DeepSeek-AI et~al.(2026)DeepSeek-AI, Xu, Lin, Xue, Wang, Xu, Wu, Zhang, Lin, Dong, Ling, Lu, Zhao, Deng, Hou, Xu, Shao, Ruan, Sun, Dai, Guo, Yang, Chen, Li, Ji, Li, Wei, Lin, Yuan, Xia, Dai, Hao, Chen, Cao, Meng, Li, Yu, Zhang, Xu, Li, Liang, Zhang, Luo, Wei, Yuan, Zhang, Luo, Chen, Ji, Zhang, Ding, Tang, Cao, Gao, Qu, Zeng, Yang, Zhu, Luo, Song, Yu, Huang, Cai, Liang, Zhou, Ye, Li, Xu, Hu, Yang, Chen, Yan, Chen, Zhou, Xiang, Yuan, Cheng, Zhou, Zhu, Yu, Sun, Ran, Jiang, Qiu, Li, Zheng, Song, Dong, Gao, Guan, Zhou, Huang, Yu, Wang, Zhang, Wang, Xia, Zhang, Zhao, Guo, Luo, Ma, Zhu, Wang, Cai, Zhang, Chen, Di, Xu, Mei, Wang, Zhang, Zhang, Tang, Li, Zhou, Han, Wang, Huang, Wang, Cong, Wang, Zhang, Wang, Zhu, Li, Chen, Du, Jiang, Tian, Xu, Lu, Xu, Ge, Zhang, Pan, Wang, Chen, Yin, Xu, Shen, Zhang, Chen, Liu, Lu, Sun, Zhou, Chen, Cai, Nie, Wu, Chen, Hu, Liu, Hu, Ma, Wang, Yu, Zhou, Pan, Yu, Zhou, Ni, Yun, Jin, Pei, Ye, Lin, Ji, Cui, Yue, Yu, Wang, Zhang, Xiao, Zeng, An, Zhao, Liu, Liang, Pang, Luo, Yao,
  Gao, Yang, Huang, Hou, Zhang, Ma, Gao, He, Wang, Wang, Bi, Liu, Wang, Chen, Zhang, Nie, Sun, Wang, Cheng, Liu, Xie, Liu, Liu, Yu, Li, Yang, Zhang, Chen, Wang, Su, Chen, Lin, Fu, Yan, Wang, Ma, Luo, Zhang, Xu, Ma, Huang, Li, Li, Xu, Zhao, Sun, Wang, Qian, Shao, Yu, Zhang, Ding, Shi, Wu, Xiong, Ma, He, Tang, Zhou, Luo, Zhong, Piao, Wang, Zhang, Chen, Tan, Wei, Ma, Liu, Yang, Guo, Wu, Wu, Li, Cheng, Ou, Xu, Li, Wang, Yang, Xu, Wu, Meng, Zou, Zha, Xiong, Chen, Lin, Cao, Wang, Zhang, Yan, Lin, Gu, Luo, You, Liu, Zhou, Zhou, Huang, Wu, Wang, Zhao, Ren, Zhang, Sha, Fu, Ju, Xu, Xie, Zhang, Gao, Hao, Gou, Ma, Yan, Shao, Huang, Chen, Wu, Ren, Wu, Li, Zhang, Xu, Wang, Qu, Gu, Zhu, Li, Zhang, Xie, Gao, Wan, Pan, and Yao]{deepseek}
DeepSeek-AI, Anyi Xu, Bangcai Lin, Bing Xue, Bingxuan Wang, Bingzheng Xu, Bochao Wu, Bowei Zhang, Chaofan Lin, Chen Dong, Chenchen Ling, Chengda Lu, Chenggang Zhao, Chengqi Deng, Chengyu Hou, Chenhao Xu, Chenze Shao, Chong Ruan, Conner Sun, Damai Dai, Daya Guo, Dejian Yang, Deli Chen, Donghao Li, Dongjie Ji, Erhang Li, Fang Wei, Fangyun Lin, Fangzhou Yuan, Feiyu Xia, Fucong Dai, Guangbo Hao, Guanting Chen, Guoai Cao, Guolai Meng, Guowei Li, Han Yu, Han Zhang, Hanwei Xu, Hao Li, Haofen Liang, Haoling Zhang, Haoming Luo, Haoran Wei, Haotian Yuan, Haowei Zhang, Haowen Luo, Haoyu Chen, Haozhe Ji, Hengqing Zhang, Honghui Ding, Hongxuan Tang, Huanqi Cao, Huazuo Gao, Hui Qu, Hui Zeng, J~Yang, JQ~Zhu, Jia Luo, Jia Song, Jia Yu, Jialiang Huang, Jialu Cai, Jian Liang, Jiangting Zhou, Jiasheng Ye, Jiashi Li, Jiaxin Xu, Jiewen Hu, Jieyu Yang, Jin Chen, Jin Yan, Jingchang Chen, Jingli Zhou, Jingting Xiang, Jingyang Yuan, Jingyuan Cheng, Jingzi Zhou, Jinhua Zhu, Jiping Yu, Joseph Sun, Jun Ran, Junguang Jiang, Junjie Qiu,
  Junlong Li, Junmin Zheng, Junxiao Song, Kai Dong, Kaige Gao, Kang Guan, Kexing Zhou, Kezhao Huang, Kuai Yu, Lean Wang, Lecong Zhang, Lei Wang, Leyi Xia, Li~Zhang, Liang Zhao, Lihua Guo, Lingxiao Luo, Linwang Ma, Linyan Zhu, Litong Wang, Liyu Cai, Liyue Zhang, Longhao Chen, MS~Di, MY~Xu, Max Mei, Miaojun Wang, Mingchuan Zhang, Minghua Zhang, Minghui Tang, Mingming Li, Mingxu Zhou, Minmin Han, Ning Wang, Panpan Huang, Panpan Wang, Peixin Cong, Peiyi Wang, Peng Zhang, Qiancheng Wang, Qihao Zhu, Qingyang Li, Qinyu Chen, Qiushi Du, Qiwei Jiang, Rui Tian, Ruifan Xu, Ruijie Lu, Ruiling Xu, Ruiqi Ge, Ruisong Zhang, Ruizhe Pan, Runji Wang, Runqian Chen, Runqiu Yin, Runxin Xu, Ruomeng Shen, Ruoyu Zhang, Ruyi Chen, SH~Liu, Shanghao Lu, Shangmian Sun, Shangyan Zhou, Shanhuang Chen, Shaofei Cai, Shaoheng Nie, Shaoqing Wu, Shaoyuan Chen, Shengding Hu, Shengyu Liu, Shiqiang Hu, Shirong Ma, Shiyu Wang, Shuiping Yu, Shunfeng Zhou, Shuting Pan, Shuying Yu, Songyang Zhou, Tao Ni, Tao Yun, Tian Jin, Tian Pei, Tian Ye, Tianle
  Lin, Tianran Ji, Tianyi Cui, Tianyuan Yue, Tingting Yu, Tun Wang, W~Zhang, WL~Xiao, Wangding Zeng, Wei An, Weilin Zhao, Wen Liu, Wenfeng Liang, Wenjie Pang, Wenjing Luo, Wenjing Yao, Wenjun Gao, Wenkai Yang, Wenlve Huang, Wenqing Hou, Wentao Zhang, Wenting Ma, Xi~Gao, Xiang He, Xiangwen Wang, Xianzu Wang, Xiao Bi, Xiaodong Liu, Xiaohan Wang, Xiaokang Chen, Xiaokang Zhang, Xiaotao Nie, Xiaowen Sun, Xiaoxiang Wang, Xin Cheng, Xin Liu, Xin Xie, Xingchao Liu, Xingchen Liu, Xingkai Yu, Xingyou Li, Xinyu Yang, Xinyu Zhang, Xu~Chen, Xuanyu Wang, Xuecheng Su, Xueyin Chen, Xuheng Lin, Xuwei Fu, YC~Yan, YQ~Wang, YW~Ma, Yanfeng Luo, Yang Zhang, Yanhong Xu, Yanru Ma, Yanwen Huang, Yao Li, Yao Li, Yao Xu, Yao Zhao, Yaofeng Sun, Yaohui Wang, Yi~Qian, Yi~Shao, Yi~Yu, Yichao Zhang, Yifan Ding, Yifan Shi, Yijia Wu, Yiliang Xiong, Yiling Ma, Ying He, Ying Tang, Ying Zhou, Yingjia Luo, Yinmin Zhong, Yishi Piao, Yisong Wang, Yixiang Zhang, Yixiao Chen, Yixuan Tan, Yixuan Wei, Yiyang Ma, Yiyuan Liu, Yonglun Yang, Yongqiang Guo,
  Yongtong Wu, Yu~Wu, YuKun Li, Yuan Cheng, Yuan Ou, Yuanfan Xu, Yuanhao Li, Yuduan Wang, Yuehan Yang, Yuer Xu, Yuhan Wu, Yuhao Meng, Yuheng Zou, Yukun Zha, Yunfan Xiong, Yupeng Chen, Yuping Lin, Yuqian Cao, Yuqian Wang, Yushun Zhang, Yuting Yan, Yutong Lin, Yuxian Gu, Yuxiang Luo, Yuxiang You, Yuxuan Liu, Yuxuan Zhou, Yuyang Zhou, Yuzhen Huang, ZF~Wu, Zehao Wang, Zehua Zhao, Zehui Ren, Zekai Zhang, Zhangli Sha, Zhe Fu, Zhe Ju, Zhean Xu, Zhenda Xie, Zhengyan Zhang, Zheren Gao, Zhewen Hao, Zhibin Gou, Zhicheng Ma, Zhigang Yan, Zhihong Shao, Zhixian Huang, Zhixuan Chen, Zhiyu Wu, Zhizhou Ren, Zhongyu Wu, Zhuoshu Li, Zhuping Zhang, Zian Xu, Zihao Wang, Zihua Qu, Zihui Gu, Zijia Zhu, Zilin Li, Zipeng Zhang, Ziwei Xie, Ziyi Gao, Ziyi Wan, Zizheng Pan, and Zongqing Yao.
\newblock Deepseek-v4: Towards highly efficient million-token context intelligence, 2026.
\newblock \url{https://arxiv.org/abs/2606.19348}.

\bibitem[Dong et~al.(2026{\natexlab{a}})Dong, Lu, Huang, Zhong, Liu, Huang, Li, Zhao, Song, Li, Jin, Zhu, Wang, Lei, Luo, Chen, Chen, Feng, Wen, and Dou]{dong2026agent}
Guanting Dong, Junting Lu, Junjie Huang, Wanjun Zhong, Longxiang Liu, Shijue Huang, Zhenyu Li, Yang Zhao, Xiaoshuai Song, Xiaoxi Li, Jiajie Jin, Yutao Zhu, Hanbin Wang, Fangyu Lei, Qinyu Luo, Mingyang Chen, Zehui Chen, Jiazhan Feng, Ji{-}Rong Wen, and Zhicheng Dou.
\newblock Agent-world: Scaling real-world environment synthesis for evolving general agent intelligence.
\newblock \emph{CoRR}, abs/2604.18292, 2026{\natexlab{a}}.
\newblock \doi{10.48550/ARXIV.2604.18292}.
\newblock \url{https://doi.org/10.48550/arXiv.2604.18292}.

\bibitem[Dong et~al.(2026{\natexlab{b}})Dong, Yang, Jin, Hu, Song, Chen, Li, Tan, Zhu, Chua, Wen, and Dou]{agentreflex}
Guanting Dong, Xinyu Yang, Jiajie Jin, Yuyang Hu, Xiaoshuai Song, Yifei Chen, Xiaoxi Li, Jiejun Tan, Yutao Zhu, Tat-Seng Chua, Ji-Rong Wen, and Zhicheng Dou.
\newblock Agent-reflex: Teaching language agents to act reflectively, 2026{\natexlab{b}}.
\newblock \url{https://arxiv.org/abs/2610.17953}.

\bibitem[Ester et~al.(1996)Ester, Kriegel, Sander, and Xu]{dbscan}
Martin Ester, Hans-Peter Kriegel, J{\"o}rg Sander, and Xiaowei Xu.
\newblock A density-based algorithm for discovering clusters in large spatial databases with noise.
\newblock In \emph{Proceedings of the ACM SIGKDD International Conference on Knowledge Discovery and Data Mining}, pages 226--231, 1996.

\bibitem[Fang et~al.(2025)Fang, Cai, Li, Wu, Li, Yin, Wang, Wang, Su, Zhang, Wu, Tao, Jiang, Xie, Huang, and Zhou]{envscaling}
Runnan Fang, Shihao Cai, Baixuan Li, Jialong Wu, Guangyu Li, Wenbiao Yin, Xinyu Wang, Xiaobin Wang, Liangcai Su, Zhen Zhang, Shibin Wu, Zhengwei Tao, Yong Jiang, Pengjun Xie, Fei Huang, and Jingren Zhou.
\newblock Towards general agentic intelligence via environment scaling, 2025.
\newblock \url{https://arxiv.org/abs/2509.13311}.

\bibitem[Guo et~al.(2025)Guo, Yang, Chen, Xiao, Wang, Juan, Qiu, Shen, and Wang]{genenv}
Jiacheng Guo, Ling Yang, Peter Chen, Qixin Xiao, Yinjie Wang, Xinzhe Juan, Jiahao Qiu, Ke~Shen, and Mengdi Wang.
\newblock Genenv: Difficulty-aligned co-evolution between llm agents and environment simulators, 2025.
\newblock \url{https://arxiv.org/abs/2512.19682}.

\bibitem[{Harvey AI}(2026)]{harveylab}
{Harvey AI}.
\newblock Harvey lab: The legal agent benchmark, 2026.
\newblock \url{https://github.com/harveyai/harvey-labs/tree/v1.0}.
\newblock Announcement: \url{https://www.harvey.ai/blog/introducing-harveys-legal-agent-benchmark}.

\bibitem[He et~al.(2026)He, Wang, Liu, Mao, Liu, Suo, Zhou, Xi, Dong, Shao, Zheng, Zhou, Wan, Gui, Zhang, Qiu, and Huang]{workrover}
Hang He, Linting Wang, Guanming Liu, Bo~Mao, Jinxiu Liu, Yiran Suo, Maosen Zhou, Zhiheng Xi, Guanting Dong, Zhou Shao, Rui Zheng, Jie Zhou, Chengcheng Wan, Tao Gui, Qi~Zhang, Xipeng Qiu, and Xuanjing Huang.
\newblock Workrover: Scaling long-horizon cowork agent training via recursive environment evolution, 2026.
\newblock \url{https://arxiv.org/abs/2610.18067}.

\bibitem[Li et~al.(2023)Li, Zhao, Yu, Song, Li, Yu, Li, Huang, and Li]{apibank}
Minghao Li, Yingxiu Zhao, Bowen Yu, Feifan Song, Hangyu Li, Haiyang Yu, Zhoujun Li, Fei Huang, and Yongbin Li.
\newblock {API}-bank: A comprehensive benchmark for tool-augmented {LLM}s.
\newblock In Houda Bouamor, Juan Pino, and Kalika Bali, editors, \emph{Proceedings of the 2023 Conference on Empirical Methods in Natural Language Processing}, pages 3102--3116, Singapore, December 2023. Association for Computational Linguistics.
\newblock \doi{10.18653/v1/2023.emnlp-main.187}.
\newblock \url{https://aclanthology.org/2023.emnlp-main.187/}.

\bibitem[Li et~al.(2026{\natexlab{a}})Li, Chen, Liu, Zheng, Chen, He, Li, You, Shen, Sun, Wang, Zeng, Wang, Zhao, Wang, Chaim, Di, Gao, He, He, Jing, Kong, Lan, Li, Li, Li, Lin, Liu, Liu, Lyu, Ma, Wang, Wang, Wang, Ye, Zhang, Xing, Xue, Dillmann, and Lee]{skillsbench}
Xiangyi Li, Wenbo Chen, Yimin Liu, Shenghan Zheng, Xiaokun Chen, Yifeng He, Yubo Li, Bingran You, Haotian Shen, Jiankai Sun, Shuyi Wang, Qunhong Zeng, Di~Wang, Xuandong Zhao, Yuanli Wang, Roey~Ben Chaim, Zonglin Di, Yipeng Gao, Junwei He, Yizhuo He, Liqiang Jing, Luyang Kong, Xin Lan, Jiachen Li, Songlin Li, Yijiang Li, Yueqian Lin, Xinyi Liu, Xuanqing Liu, Haoran Lyu, Ze~Ma, Bowei Wang, Runhui Wang, Tianfu Wang, Wengao Ye, Yue Zhang, Hanwen Xing, Yiqi Xue, Steven Dillmann, and Han{-}Chung Lee.
\newblock Skillsbench: Benchmarking how well agent skills work across diverse tasks.
\newblock \emph{CoRR}, abs/2602.12670, 2026{\natexlab{a}}.
\newblock \url{https://arxiv.org/abs/2602.12670}.

\bibitem[Li et~al.(2025)Li, Inan, Yue, Chen, Wutschitz, Kulkarni, Poovendran, Sim, and Rajmohan]{simia}
Yuetai Li, Huseyin~A Inan, Xiang Yue, Wei-Ning Chen, Lukas Wutschitz, Janardhan Kulkarni, Radha Poovendran, Robert Sim, and Saravan Rajmohan.
\newblock Simulating environments with reasoning models for agent training, 2025.
\newblock \url{https://arxiv.org/abs/2511.01824}.

\bibitem[Li et~al.(2026{\natexlab{b}})Li, Feng, Xu, Ma, Zheng, Jiang, Sun, Shao, Chen, Huang, Han, Lee, Xu, Zeng, Hua, Zhang, Alomair, Krishna, Zettlemoyer, Koh, Ramasubramanian, Niu, Yue, and Poovendran]{jobbench}
Yuetai Li, Yichen Feng, Zhangchen Xu, Zixian Ma, Kaiyuan Zheng, Fengqing Jiang, Xinghua Sun, Rulin Shao, Zichen Chen, Yue Huang, Xinyang Han, Brian Lee, Kayla Xu, Shenglai Zeng, Hang Hua, Xiangliang Zhang, Basel Alomair, Ranjay Krishna, Luke Zettlemoyer, Pang~Wei Koh, Bhaskar Ramasubramanian, Luyao Niu, Xiang Yue, and Radha Poovendran.
\newblock Jobbench: Aligning agent work with human will, 2026{\natexlab{b}}.
\newblock \url{https://arxiv.org/abs/2605.26329}.

\bibitem[Long et~al.(2026)Long, Chen, Chen, Zhang, Hu, Bao, and Wang]{trace2env}
Quanyu Long, Xiao Chen, Jianda Chen, Haozhen Zhang, Qisheng Hu, Jianzhu Bao, and Wenya Wang.
\newblock From traces to agentic worlds: Agentic language world models for interactive environment simulation.
\newblock \emph{CoRR}, abs/2610.06100, 2026.
\newblock \url{https://doi.org/10.48550/arXiv.2610.06100}.

\bibitem[Mao et~al.(2026)Mao, He, Wang, Lin, Zhou, Liu, Liu, Huai, Zhang, Li, Lei, Dong, Shao, Zheng, Yan, Zhou, Wan, Gui, He, and Qiu]{weft}
Bo~Mao, Hang He, Linting Wang, Lizhi Lin, Maosen Zhou, Guanming Liu, Jinxiu Liu, Tianyu Huai, Chaoyun Zhang, Bingxuan Li, Kepeng Lei, Guanting Dong, Zhou Shao, Rui Zheng, Hang Yan, Jie Zhou, Chengcheng Wan, Tao Gui, Liang He, and Xipeng Qiu.
\newblock Weft: Scaling tool-use post-training for general-purpose agents, 2026.
\newblock \url{https://arxiv.org/abs/2609.36887}.

\bibitem[{Meta}(2026)]{museglimmer}
{Meta}.
\newblock {Muse Glimmer 30B}: An on-device agentic model.
\newblock \url{https://huggingface.co/meta-models/Muse-Glimmer-30B}, 2026.
\newblock Official model card; accessed 26 Oct. 2026.

\bibitem[{Moonshot AI}(2026)]{kimi}
{Moonshot AI}.
\newblock {Kimi K3}.
\newblock \url{https://huggingface.co/moonshotai/Kimi-K3}, 2026.
\newblock Official model card and technical report; accessed 26 Oct. 2026.

\bibitem[{Nex-AGI}(2026)]{nex}
{Nex-AGI}.
\newblock {Nex-N2-mini}.
\newblock \url{https://huggingface.co/nex-agi/Nex-N2-mini}, 2026.
\newblock Official model card; accessed 26 Oct. 2026.

\bibitem[{OpenAI}(2026{\natexlab{a}})]{openai}
{OpenAI}.
\newblock {GPT-5.6}: Frontier intelligence that scales with your ambition.
\newblock \url{https://openai.com/index/gpt-5-6/}, 2026{\natexlab{a}}.
\newblock Official announcement of the GPT-5.6 family incl. GPT-5.6 Sol, 9 Jul. 2026; accessed 26 Oct. 2026.

\bibitem[{OpenAI}(2026{\natexlab{b}})]{openai2026gpt61sol}
{OpenAI}.
\newblock Addendum: Gpt-6.1 sol system card.
\newblock \url{https://cdn.openai.com/pdf/38e3efcf-545e-44cd-99ec-2b7eb395f4cc/oai_GPT_6_1_Sol.pdf}, September 2026{\natexlab{b}}.
\newblock Released September 29, 2026.

\bibitem[Opsahl{-}Ong et~al.(2026)Opsahl{-}Ong, Singhvi, Collins, Zhou, Wang, Baheti, Oertell, Portes, Havens, Elsen, Bendersky, Zaharia, and Chen]{officeqa}
Krista Opsahl{-}Ong, Arnav Singhvi, Jasmine Collins, Ivan Zhou, Cindy Wang, Ashutosh Baheti, Owen Oertell, Jacob Portes, Sam Havens, Erich Elsen, Michael Bendersky, Matei Zaharia, and Xing Chen.
\newblock Officeqa pro: An enterprise benchmark for end-to-end grounded reasoning.
\newblock \emph{CoRR}, abs/2603.08655, 2026.
\newblock \url{https://arxiv.org/abs/2603.08655}.

\bibitem[Patil et~al.(2023)Patil, Zhang, Wang, and Gonzalez]{gorilla}
Shishir~G. Patil, Tianjun Zhang, Xin Wang, and Joseph~E. Gonzalez.
\newblock Gorilla: Large language model connected with massive apis, 2023.
\newblock \url{https://arxiv.org/abs/2305.15334}.

\bibitem[Patil et~al.(2025)Patil, Mao, Yan, Ji, Suresh, Stoica, and Gonzalez]{bfcl}
Shishir~G. Patil, Huanzhi Mao, Fanjia Yan, Charlie Cheng-Jie Ji, Vishnu Suresh, Ion Stoica, and Joseph~E. Gonzalez.
\newblock The berkeley function calling leaderboard (bfcl): From tool use to agentic evaluation of large language models.
\newblock In \emph{Proceedings of the 42nd International Conference on Machine Learning}, volume 267 of \emph{Proceedings of Machine Learning Research}, pages 48371--48392. PMLR, 2025.
\newblock \url{https://proceedings.mlr.press/v267/patil25a.html}.

\bibitem[Prabhakar et~al.(2025)Prabhakar, Liu, Zhu, Zhang, Awalgaonkar, Wang, Liu, Chen, Hoang, Niebles, Heinecke, Yao, Wang, Savarese, and Xiong]{apigenmt}
Akshara Prabhakar, Zuxin Liu, Ming Zhu, Jianguo Zhang, Tulika Awalgaonkar, Shiyu Wang, Zhiwei Liu, Haolin Chen, Thai Hoang, Juan~Carlos Niebles, Shelby Heinecke, Weiran Yao, Huan Wang, Silvio Savarese, and Caiming Xiong.
\newblock Apigen-mt: Agentic pipeline for multi-turn data generation via simulated agent-human interplay, 2025.
\newblock \url{https://arxiv.org/abs/2504.03601}.

\bibitem[Qin et~al.(2023)Qin, Liang, Ye, Zhu, Yan, Lu, Lin, Cong, Tang, Qian, Zhao, Hong, Tian, Xie, Zhou, Gerstein, Li, Liu, and Sun]{toolllm}
Yujia Qin, Shihao Liang, Yining Ye, Kunlun Zhu, Lan Yan, Yaxi Lu, Yankai Lin, Xin Cong, Xiangru Tang, Bill Qian, Sihan Zhao, Lauren Hong, Runchu Tian, Ruobing Xie, Jie Zhou, Mark Gerstein, Dahai Li, Zhiyuan Liu, and Maosong Sun.
\newblock Toolllm: Facilitating large language models to master 16000+ real-world apis, 2023.
\newblock \url{https://arxiv.org/abs/2307.16789}.

\bibitem[{Qwen Team}(2026{\natexlab{a}})]{qwen3.6-27b}
{Qwen Team}.
\newblock {Qwen3.6-27B}: Flagship-level coding in a {27B} dense model, April 2026{\natexlab{a}}.
\newblock \url{https://qwen.ai/blog?id=qwen3.6-27b}.

\bibitem[{Qwen Team}(2026{\natexlab{b}})]{qwen36}
{Qwen Team}.
\newblock {Qwen3.6-35B-A3B}.
\newblock \url{https://huggingface.co/Qwen/Qwen3.6-35B-A3B}, 2026{\natexlab{b}}.
\newblock Official model card; accessed 26 Oct. 2026.

\bibitem[{Qwen Team}(2026{\natexlab{c}})]{qwen38max}
{Qwen Team}.
\newblock {Qwen3.8-Max}: A new bar for coding and cowork.
\newblock \url{https://qwen.ai/blog?id=qwen3.8}, 2026{\natexlab{c}}.
\newblock Official release post; accessed 26 Oct. 2026.

\bibitem[Schick et~al.(2023)Schick, Dwivedi-Yu, Dessi, Raileanu, Lomeli, Hambro, Zettlemoyer, Cancedda, and Scialom]{toolformer}
Timo Schick, Jane Dwivedi-Yu, Roberto Dessi, Roberta Raileanu, Maria Lomeli, Eric Hambro, Luke Zettlemoyer, Nicola Cancedda, and Thomas Scialom.
\newblock Toolformer: Language models can teach themselves to use tools.
\newblock In A.~Oh, T.~Naumann, A.~Globerson, K.~Saenko, M.~Hardt, and S.~Levine, editors, \emph{Advances in Neural Information Processing Systems}, volume~36, pages 68539--68551. Curran Associates, Inc., 2023.
\newblock \doi{10.52202/075280-2997}.
\newblock \url{https://proceedings.neurips.cc/paper_files/paper/2023/file/d842425e4bf79ba039352da0f658a906-Paper-Conference.pdf}.

\bibitem[Shao et~al.(2024)Shao, Wang, Zhu, Xu, Song, Bi, Zhang, Zhang, Li, Wu, and Guo]{shao2024deepseekmath}
Zhihong Shao, Peiyi Wang, Qihao Zhu, Runxin Xu, Junxiao Song, Xiao Bi, Haowei Zhang, Mingchuan Zhang, Y.~K. Li, Y.~Wu, and Daya Guo.
\newblock Deepseekmath: Pushing the limits of mathematical reasoning in open language models, 2024.
\newblock \url{https://arxiv.org/abs/2402.03300}.

\bibitem[Shepard and Salimans(2026)]{automationbench}
Daniel Shepard and Robin Salimans.
\newblock Automationbench, 2026.
\newblock \url{https://arxiv.org/abs/2604.18934}.

\bibitem[Shi et~al.(2026)Shi, Zytek, Razavi, Narasimhan, and Barres]{tauknowledge}
Quan Shi, Alexandra Zytek, Pedram Razavi, Karthik~R Narasimhan, and Victor Barres.
\newblock $\tau$-knowledge: Evaluating conversational agents over unstructured knowledge.
\newblock In Tong Zhang, Miroslav Dudik, Martin Jaggi, Alekh Agarwal, Sharon Li, Dale Schuurmans, Jerry Zhu, Felix Berkenkamp, Hanze Dong, and Alberto Bietti, editors, \emph{Proceedings of the 43rd International Conference on Machine Learning}, volume 306 of \emph{Proceedings of Machine Learning Research}, pages 112443--112471. PMLR, 06--11 Jul 2026.
\newblock \url{https://proceedings.mlr.press/v306/shi26ah.html}.

\bibitem[Shinn et~al.(2023)Shinn, Cassano, Gopinath, Narasimhan, and Yao]{reflexion}
Noah Shinn, Federico Cassano, Ashwin Gopinath, Karthik Narasimhan, and Shunyu Yao.
\newblock Reflexion: language agents with verbal reinforcement learning.
\newblock In A.~Oh, T.~Naumann, A.~Globerson, K.~Saenko, M.~Hardt, and S.~Levine, editors, \emph{Advances in Neural Information Processing Systems}, volume~36, pages 8634--8652. Curran Associates, Inc., 2023.
\newblock \doi{10.52202/075280-0377}.
\newblock \url{https://proceedings.neurips.cc/paper_files/paper/2023/file/1b44b878bb782e6954cd888628510e90-Paper-Conference.pdf}.

\bibitem[Song et~al.(2026)Song, Chang, Dong, Zhu, Wen, and Dou]{envscaler}
Xiaoshuai Song, Haofei Chang, Guanting Dong, Yutao Zhu, Ji-Rong Wen, and Zhicheng Dou.
\newblock Envscaler: Scaling tool-interactive environments for llm agent via programmatic synthesis, 2026.
\newblock \url{https://arxiv.org/abs/2601.05808}.

\bibitem[Song et~al.(2023)Song, Xiong, Zhu, Wu, Qian, Song, Huang, Li, Wang, Yao, Tian, and Li]{restgpt}
Yifan Song, Weimin Xiong, Dawei Zhu, Wenhao Wu, Han Qian, Mingbo Song, Hailiang Huang, Cheng Li, Ke~Wang, Rong Yao, Ye~Tian, and Sujian Li.
\newblock Restgpt: Connecting large language models with real-world restful apis, 2023.
\newblock \url{https://arxiv.org/abs/2306.06624}.

\bibitem[Tao et~al.(2025)Tao, Wu, Yin, Zhang, Li, Shen, Li, Zhang, Wang, Jiang, Xie, Huang, and Zhou]{webshaper}
Zhengwei Tao, Jialong Wu, Wenbiao Yin, Junkai Zhang, Baixuan Li, Haiyang Shen, Kuan Li, Liwen Zhang, Xinyu Wang, Yong Jiang, Pengjun Xie, Fei Huang, and Jingren Zhou.
\newblock Webshaper: Agentically data synthesizing via information-seeking formalization, 2025.
\newblock \url{https://arxiv.org/abs/2507.15061}.

\bibitem[Wang et~al.(2025)Wang, Li, Song, Xu, Tang, Zhuge, Pan, Song, Li, Singh, Tran, Li, Ma, Zheng, Qian, Shao, Muennighoff, Zhang, Hui, Lin, Brennan, Peng, Ji, and Neubig]{openhands}
Xingyao Wang, Boxuan Li, Yufan Song, Frank~F. Xu, Xiangru Tang, Mingchen Zhuge, Jiayi Pan, Yueqi Song, Bowen Li, Jaskirat Singh, Hoang~H. Tran, Fuqiang Li, Ren Ma, Mingzhang Zheng, Bill Qian, Yanjun Shao, Niklas Muennighoff, Yizhe Zhang, Binyuan Hui, Junyang Lin, Robert Brennan, Hao Peng, Heng Ji, and Graham Neubig.
\newblock Openhands: An open platform for ai software developers as generalist agents, 2025.
\newblock \url{https://arxiv.org/abs/2407.16741}.

\bibitem[Wang et~al.(2026)Wang, Xu, Liu, Wang, Han, Yao, Yao, and He]{awm}
Zhaoyang Wang, Canwen Xu, Boyi Liu, Yite Wang, Siwei Han, Zhewei Yao, Huaxiu Yao, and Yuxiong He.
\newblock Agent world model: Infinity synthetic environments for agentic reinforcement learning, 2026.
\newblock \url{https://arxiv.org/abs/2602.10090}.

\bibitem[Wu et~al.(2025)Wu, Li, Fang, Yin, Zhang, Tao, Zhang, Xi, Fu, Jiang, Xie, Huang, and Zhou]{webdancer}
Jialong Wu, Baixuan Li, Runnan Fang, Wenbiao Yin, Liwen Zhang, Zhengwei Tao, Dingchu Zhang, Zekun Xi, Gang Fu, Yong Jiang, Pengjun Xie, Fei Huang, and Jingren Zhou.
\newblock Webdancer: Towards autonomous information seeking agency, 2025.
\newblock \url{https://arxiv.org/abs/2505.22648}.

\bibitem[Wu et~al.(2026)Wu, Zhang, Zhang, Wang, Su, Chen, Wang, Wang, Shen, Zhou, Yang, Huang, Yang, and Liu]{terminaluniverse}
Jie Wu, Zhenru Zhang, Beichen Zhang, Xuwu Wang, Yuhui Su, Mouxiang Chen, Peng Wang, Zhihai Wang, Que Shen, Hao Zhou, An~Yang, Fei Huang, Yujiu Yang, and Dayiheng Liu.
\newblock Terminal-universe: Turning agent trajectories into scalable terminal environments.
\newblock \emph{CoRR}, abs/2609.04148, 2026.
\newblock \url{https://doi.org/10.48550/arXiv.2609.04148}.

\bibitem[Xu et~al.(2025)Xu, Soria, Tan, Roy, Agrawal, Poovendran, and Panda]{toucan}
Zhangchen Xu, Adriana~Meza Soria, Shawn Tan, Anurag Roy, Ashish~Sunil Agrawal, Radha Poovendran, and Rameswar Panda.
\newblock Toucan: Synthesizing 1.5m tool-agentic data from real-world mcp environments, 2025.
\newblock \url{https://arxiv.org/abs/2510.01179}.

\bibitem[Yang et~al.(2025)Yang, Li, Yang, Zhang, Hui, Zheng, Yu, Gao, Huang, Lv, Zheng, Liu, Zhou, Huang, Hu, Ge, Wei, Lin, Tang, Yang, Tu, Zhang, Yang, Yang, Zhou, Zhou, Lin, Dang, Bao, Yang, Yu, Deng, Li, Xue, Li, Zhang, Wang, Zhu, Men, Gao, Liu, Luo, Li, Tang, Yin, Ren, Wang, Zhang, Ren, Fan, Su, Zhang, Zhang, Wan, Liu, Wang, Cui, Zhang, Zhou, and Qiu]{qwen3}
An~Yang, Anfeng Li, Baosong Yang, Beichen Zhang, Binyuan Hui, Bo~Zheng, Bowen Yu, Chang Gao, Chengen Huang, Chenxu Lv, Chujie Zheng, Dayiheng Liu, Fan Zhou, Fei Huang, Feng Hu, Hao Ge, Haoran Wei, Huan Lin, Jialong Tang, Jian Yang, Jianhong Tu, Jianwei Zhang, Jianxin Yang, Jiaxi Yang, Jing Zhou, Jingren Zhou, Junyang Lin, Kai Dang, Keqin Bao, Kexin Yang, Le~Yu, Lianghao Deng, Mei Li, Mingfeng Xue, Mingze Li, Pei Zhang, Peng Wang, Qin Zhu, Rui Men, Ruize Gao, Shixuan Liu, Shuang Luo, Tianhao Li, Tianyi Tang, Wenbiao Yin, Xingzhang Ren, Xinyu Wang, Xinyu Zhang, Xuancheng Ren, Yang Fan, Yang Su, Yichang Zhang, Yinger Zhang, Yu~Wan, Yuqiong Liu, Zekun Wang, Zeyu Cui, Zhenru Zhang, Zhipeng Zhou, and Zihan Qiu.
\newblock Qwen3 technical report, 2025.
\newblock \url{https://arxiv.org/abs/2505.09388}.

\bibitem[Yang et~al.(2024)Yang, Jimenez, Wettig, Lieret, Yao, Narasimhan, and Press]{sweagent}
John Yang, Carlos Jimenez, Alexander Wettig, Kilian Lieret, Shunyu Yao, Karthik Narasimhan, and Ofir Press.
\newblock Swe-agent: Agent-computer interfaces enable automated software engineering.
\newblock In A.~Globerson, L.~Mackey, D.~Belgrave, A.~Fan, U.~Paquet, J.~Tomczak, and C.~Zhang, editors, \emph{Advances in Neural Information Processing Systems}, volume~37, pages 50528--50652. Curran Associates, Inc., 2024.
\newblock \doi{10.52202/079017-1601}.
\newblock \url{https://proceedings.neurips.cc/paper_files/paper/2024/file/5a7c947568c1b1328ccc5230172e1e7c-Paper-Conference.pdf}.

\bibitem[Yao et~al.(2023)Yao, Zhao, Yu, Du, Shafran, Narasimhan, and Cao]{react}
Shunyu Yao, Jeffrey Zhao, Dian Yu, Nan Du, Izhak Shafran, Karthik Narasimhan, and Yuan Cao.
\newblock React: Synergizing reasoning and acting in language models, 2023.
\newblock \url{https://arxiv.org/abs/2210.03629}.

\bibitem[Yin et~al.(2025)Yin, Wang, Hsu, Yan, Jiang, Chen, Gu, Le, Chang, Lee, Palangi, and Pfister]{magnet}
Fan Yin, Zifeng Wang, I-Hung Hsu, Jun Yan, Ke~Jiang, Yanfei Chen, Jindong Gu, Long Le, Kai-Wei Chang, Chen-Yu Lee, Hamid Palangi, and Tomas Pfister.
\newblock Magnet: Multi-turn tool-use data synthesis and distillation via graph translation.
\newblock In Wanxiang Che, Joyce Nabende, Ekaterina Shutova, and Mohammad~Taher Pilehvar, editors, \emph{Proceedings of the 63rd Annual Meeting of the {A}ssociation for {C}omputational {L}inguistics (Volume 1: Long Papers)}, pages 32600--32616, Vienna, Austria, July 2025. Association for Computational Linguistics.
\newblock ISBN 979-8-89176-251-0.
\newblock \doi{10.18653/v1/2025.acl-long.1566}.
\newblock \url{https://aclanthology.org/2025.acl-long.1566/}.

\bibitem[Zhang et~al.(2025)Zhang, Shang, Gao, Zhang, Xie, Ma, Yuan, Wu, Zhu, and Li]{tongui}
Bofei Zhang, Zirui Shang, Zhi Gao, Wang Zhang, Rui Xie, Xiaojian Ma, Tao Yuan, Xinxiao Wu, Song-Chun Zhu, and Qing Li.
\newblock Tongui: Internet-scale trajectories from multimodal web tutorials for generalized gui agents, 2025.
\newblock \url{https://arxiv.org/abs/2504.12679}.

\bibitem[{Zhipu AI}(2026{\natexlab{a}})]{glm52}
{Zhipu AI}.
\newblock {GLM-5.2}: Built for long-horizon tasks.
\newblock \url{https://huggingface.co/zai-org/GLM-5.2}, 2026{\natexlab{a}}.
\newblock Official model card; accessed 26 Oct. 2026.

\bibitem[{Zhipu AI}(2026{\natexlab{b}})]{glm53}
{Zhipu AI}.
\newblock {GLM-5.3}: Frontier coding with emergent cyber capabilities.
\newblock \url{https://z.ai/blog/glm-5.3}, 2026{\natexlab{b}}.
\newblock Official release post, 14 Aug. 2026; accessed 26 Oct. 2026.

\end{thebibliography}

\end{document}